\documentclass[lettersize,journal]{IEEEtran}
\usepackage{amsmath,amsfonts}
\usepackage{algorithmic}
\usepackage{algorithm}
\usepackage{array}
\usepackage{amsmath}
\usepackage{amssymb}
\usepackage{mathtools}
\usepackage{amsthm}
\usepackage[subfigure]{graphfig}
\usepackage{cases} 
\usepackage{stmaryrd}                                
\usepackage{psfrag}
\usepackage{placeins}                                
\usepackage{colortbl}                                
\usepackage{amsthm}                                  
\usepackage{multirow,booktabs}
\usepackage[table,xcdraw]{xcolor}
\usepackage[normalem]{ulem}
\useunder{\uline}{\ul}{}
\usepackage{footnote}
\usepackage{epstopdf}
\usepackage{array}
\usepackage{ulem}
\usepackage{float}
\usepackage{ragged2e}
\usepackage{hyperref}
\usepackage{enumitem}
\usepackage{pifont}
\begin{document}

\title{Mask-informed Deep Contrastive Incomplete Multi-view Clustering}

\author{Zhenglai Li, ~\IEEEmembership{Member,~IEEE,}
	Yuqi Shi,
	Xiao He,
	Chang Tang, ~\IEEEmembership{Senior Member,~IEEE,}
\thanks{Z. Li is with the Faculty of Data Science, City University of Macau, Macau 999078, China, and also with the Shenzhen Institute of Advanced Technology, Chinese Academy of Sciences, Shenzhen 518055, China.
(E-mail: zhenglai.li@siat.ac.cn).}
\thanks{Y. Shi is with the Faculty of Science and Technology (The State Key Laboratory of Internet of Things for Smart City), University of Macau, 999078, China.
(E-mail:   shiyuqi@stu.ouc.edu.cn).}
\thanks{X. He and C. Tang are with the School of Computer Science, China University of Geosciences, Wuhan 430074, China.
(E-mail:\{xiaoh,tangchang\}@cug.edu.cn).}
\thanks{Manuscript received xx xx, xx; revised xx xx, xx. Zhenglai Li and Yuqi Shi contribute equally to this work. The work was supported in part by the National Natural Science Foundation of China under grant 62476258 and 62325604. (Corresponding author: Chang Tang.)}}

\markboth{Journal of \LaTeX\ Class Files,~Vol.~14, No.~8, August~2021}%
{Shell \MakeLowercase{\textit{et al.}}: A Sample Article Using IEEEtran.cls for IEEE Journals}


\maketitle

\begin{abstract}
	Multi-view clustering (MvC) utilizes information from multiple views to uncover the underlying structures of data. Despite significant advancements in MvC, mitigating the impact of missing samples in specific views on the integration of knowledge from different views remains a critical challenge. This paper proposes a novel Mask-informed Deep Contrastive Incomplete Multi-view Clustering (Mask-IMvC) method, which elegantly identifies a view-common representation for clustering. Specifically, we introduce a mask-informed fusion network that aggregates incomplete multi-view information while considering the observation status of samples across various views as a mask, thereby reducing the adverse effects of missing values. Additionally, we design a prior knowledge-assisted contrastive learning loss that boosts the representation capability of the aggregated view-common representation by injecting neighborhood information of samples from different views. Finally, extensive experiments are conducted to demonstrate the superiority of the proposed Mask-IMvC method over state-of-the-art approaches across multiple MvC datasets, both in complete and incomplete scenarios. The demo code for our work will be publicly available at \url{https://github.com/guanyuezhen/Mask-IMvC}.	
\end{abstract}


\section{Introduction}

\begin{figure}[t]
	\centering
	\includegraphics[width = \linewidth]{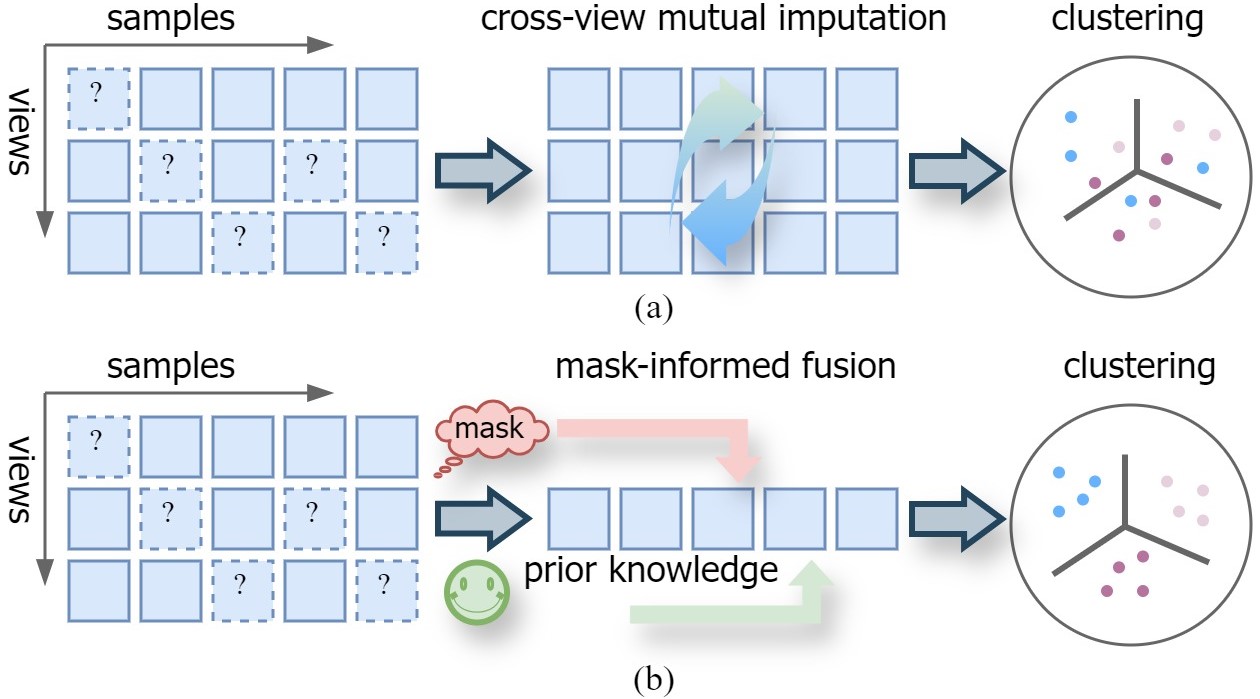}
	\caption{Comparison of two different IMvC paradigms. (a) The imputation-based IMvC. (b) Our proposed mask-informed deep contrastive IMvC.}
	\label{fig:intro}
\end{figure}

Multi-view data commonly appears in many real-world applications with the rapid development of diverse feature descriptors and data collectors~\cite{zhou2024survey, jiang2022tensorial, jia2021multi}. For instance, the visual, text, and audio information are extracted and exploited in various video understanding tasks~\cite{li2024mvbench, schiappa2023self}. To handle this type of data, multi-view learning methods are proposed to integrate the information from various views to achieve compatible performance than the approaches individually trained on a single view~\cite{li2024mvbench}. Multi-view clustering (MvC), one of the kinds of multi-view learning methods, groups the data points into their respective clusters by exploring the comprehensive information among various views without labels, and has received much attention in recent years~\cite{ke2023clustering, zhao2025scalable, yan2024deep}.

In many real-world applications, the samples in particular views may be absent due to hardware faults or interference signals during information collection, resulting in incomplete multi-view data~\cite{wen2023graph}. The incomplete multi-view clustering (IMvC) methods are introduced to address this type of data and have witnessed great progress recently~\cite{wen2023graph, lin2021completer}. Previous IMvC approaches mainly perform cross-view imputation to recover the absent information among different views based on cross-view semantic consistency as illustrated in Fig.~\ref{fig:intro} (a). For example, Li et al.~\cite{li2023incomplete} introduced a prototype-based imputation manner that utilizes a dual-attention equipped with a contrastive loss to seek prototypes of each view for recovering the missing views.  Pu et al.~\cite{pu2024adaptive} designed an adaptive imputation layer that leverages the soft cluster assignments across different views and the global cluster centroids to impute the missing views in a latent embedding space. However, the imputation procedure of missing views without the guidance of true data distributions inevitably introduces noise information and consequentially reduces the quality of imputed views. As a result, the low-quality imputed views further degrade the learning process of the IMvC methods, leading to an under-optimal view-common representation for clustering.

Different from previous methods, we propose to cluster the incomplete multi-view data without the missing view imputation to avoid introducing inaccurate information. To achieve this goal, the following challenge needs to be addressed: \textit{How can we reduce the impact of missing values across different views when merging information from multiple sources while learning a discriminative view-common feature representation for data clustering?} The missing views are difficult to be directly fused due the missing values. Even some imputation manners are exploited to handle the missing views so that the multi-view features are easier to be aggregated, the uncontrollable information underlying the missing views contains inevitable adverse effects during the fusion process. Thus, it is emerging to design a flexible and effective multi-view fusion strategy to learn and integrate multi-view information when facing either complete or incomplete parts of the data. The final target of multi-view information fusion is to seek a view-common representation for clustering. The prior knowledge, e.g. the manifold structures~\cite{meilua2024manifold} of different views, is valuable for representation learning since the prior knowledge helps to regularize the aggregated view-common feature representation with a compact and discriminative structure that polishes the noise information. Yet previous methods usually ignore utilizing such prior structural knowledge to boost the aggregated view-common features in the IMvC task.

Based on the above discussion, we propose Mask-IMvC (short for Mask-informed Deep Contrastive Incomplete Multi-view Clustering), a novel solution that flexibly and effectively integrates the information among incomplete multi-view data for clustering. As shown in Fig~\ref{fig:intro} (b), we first introduce a mask-informed fusion network that takes the observation status of samples across various views into consideration to formulate a mask, removing the contributions of missing sample during the multi-view information aggregation. Afterward, a prior knowledge-assisted contrastive learning is designed, in which the neighbor correlations of each sample captured from different views are injected into the learned view-common representation with a re-weighted contrastive loss to achieve a more compact and clear cluster structure. As s result, the incomplete multi-view information is elegantly aggregated into a view-common feature representation for clustering.
The contributions are summarized as follows,
\begin{itemize}
	\item We propose Mask-IMvC which is a flexible and effective imputation-free framework to integrate comprehensive information from incomplete multi-view data for clustering while reducing the adverse effects of missing values.
	\item We employ a re-weighted contrastive loss to regularize the view-common representation with a compact and clear cluster structure with the guidance of neighbor connection probabilities among samples across diverse views so as to boost the clustering performance.
	\item We perform extensive experiments on various benchmark multi-view datasets to verify the effectiveness of the Mask-IMvC method on both MvC and IMvC tasks.
\end{itemize}

\section{Related Work}\label{sec:sec2}

\subsection{Incomplete Multi-view Clustering}

Existing IMvC approaches can be generally grouped into: traditional IMvC methods~\cite{wen2020adaptive, yin2021incomplete, wang2022highly} and deep IMvC algorithms~\cite{yang2022robust, xu2024deep, yan2024deep}. 

Traditional IMvC methods commonly build upon matrix factorization, kernel learning, and graph learning technologies. Matrix factorization-based approaches derive a view-shared representation across diverse views via the matrix factorization techniques~\cite{hu2019one, wen2023graph, khan2024weighted}. For instance, an adaptive feature weighting manner is inserted into matrix factorization to mitigate the effects of redundant and noisy features on the learned view-shared representation, which is also regularized with a graph-embedded consensus constraint to preserve the structural information inherent in incomplete multi-view data. Kernel-based methods~\cite{9556554, LI2024102086} utilizes a set of per-computed kernels to measure diverse views, tending to formulate a unified kernel through linear or non-linear combinations of predefined kernels to capture the clustering results. Liu et al.~\cite{9556554} jointly conduct incomplete kernel matrices imputation and alignment to capture an advanced clustering representation. Graph-based methods~\cite{yang2024geometric, du2024fast, li2022refining} integrate graph similarities captured from various views via either self-representation~\cite{elhamifar2013sparse} or adaptive neighbor graph learning~\cite{nie2014clustering} manner, ultimately obtaining the clustering results through spectral clustering. Li et al.~\cite{li2022refining} explored the cross-view information to refine the graph structure by leveraging the tensor nuclear norm. Deep learning-based methods~\cite{wen2020dimc, xu2022deep, li2023incomplete} leverage the powerful representation capabilities of deep neural networks to derive consensus clustering results from multi-view data. For example, Lin et al.~\cite{lin2021completer} learned the consensus representation across diverse views by contrastive learning and recovered the missing views by cross-view prediction. Xue et al.~\cite{xue2024robust} developed a multi-graph contrastive regularization to reduce abundant correlations
across multiple views and learn discriminative representations for clustering. 

Previous IMvC methods have made significant strides in improving clustering performance following the main pipeline that imputes missing views and conducts clustering. However, the imputation procedure without the true data distributions inevitably causes inaccurate imputed missing views, which in turn degrade clustering performance. Different from them, we learn a view-common representation only from the observed parts of incomplete multi-view without the imputation procedure. To this end, we propose a mask-informed deep contrastive incomplete multi-view clustering method that reduces the impacts of missing values among different views on formulating a view-common representation from multiple views for clustering.

\subsection{Contrastive Learning}

Contrastive learning is a novel self-supervised learning paradigm that has achieved significant success across various computer vision and machine learning tasks~\cite{krishnan2022self, liu2021self}. Its core principle involves pushing samples away from their negative anchors while pulling samples closer to their positive anchors, thereby creating a discriminative representation for downstream tasks~\cite{khosla2020supervised}. Over the past few years, different contrastive learning approaches have emerged, including MoCo~\cite{he2020momentum}, SimCLR~\cite{chen2020simple}, and SwAV~\cite{caron2020unsupervised}. For a comprehensive overview of additional methods, we refer to the survey in~\cite{gui2024survey}. 

Recently, inspired by the robust feature learning capabilities of self-supervised learning, contrastive loss has been extensively applied in MvC. For instance, Xu et al.\cite{xu2022multi} aligned multi-view information from both high-level semantics and low-level features through contrastive learning, effectively capturing common semantics for clustering. Additionally, the work in\cite{trosten2023effects} explored the effectiveness of self-supervision and contrastive alignment within the multi-view clustering task. Luo et al.~\cite{luo2024simple} first fused multi-view information at the data level and then applied data augmentation for the fused data, making a shared feature extractor with a simple contrastive learning paradigm to capture robust features for clustering. However, previous methods typically treat the same samples from different views as positive pairs, while diverse samples across multiple views are considered negative pairs. This manner lacks the flexibility to leverage sample connection probabilities to enhance feature representation learning. Thus, we propose prior knowledge-assisted contrastive learning to inject the sample connection probabilities from diverse views into the view-common feature representation with a carefully designed re-weighted contrastive loss. In such a manner, the view-common feature representation full of structural information is beneficial for recovering the underlying cluster structure of data.

\begin{figure*}[!htbp]
	\centering
	\includegraphics[width = 0.8\linewidth]{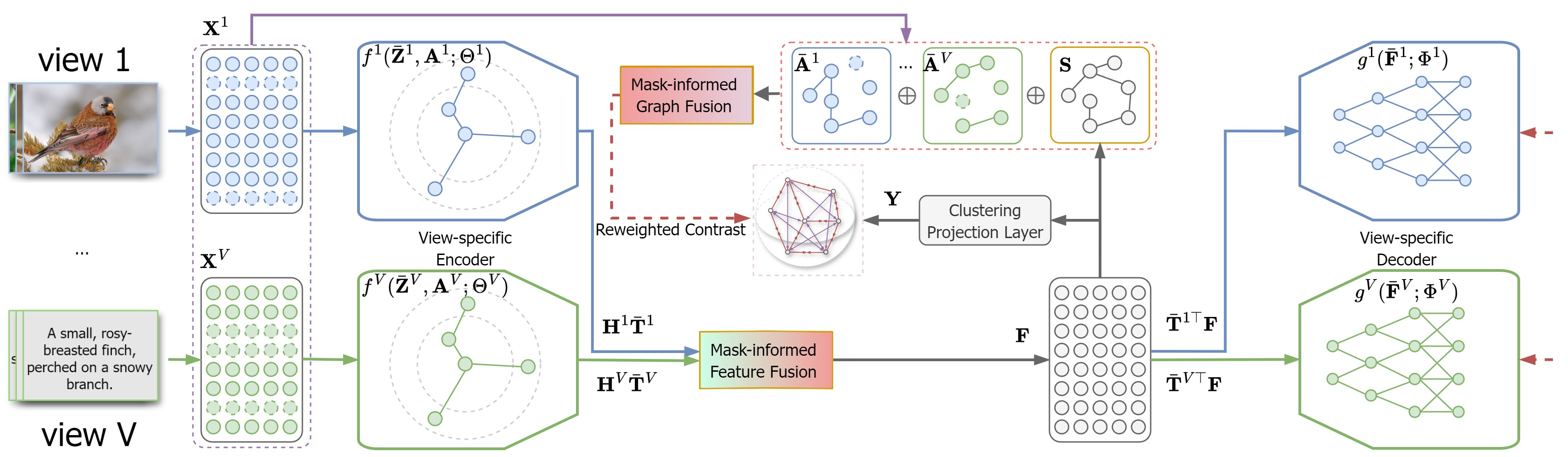}
	\caption{Illustration of mask-informed deep contrastive incomplete multi-view clustering (Mask-IMvC). The view-complete parts of IMvC data are first processed through their encoders to extract view-specific latent features. Next, a mask-informed fusion module aggregates the representations into a unified view-common one, which is then used to reconstruct the view complete parts of IMvC data via view-specific decoders. Finally, the prior knowledge from different views is fused via the mask-informed fusion strategy to assist the contrastive learning on the view-common representation.}
	\label{fig:IMvC}
\end{figure*}

\section{Methodology}

In this section, we first provide the notations used in this paper. Then, we present the details of our proposed mask-informed deep contrastive incomplete multi-view clustering framework and prior knowledge-assisted contrastive learning.

\subsection{Notation}
In this paper, the IMvC data is denoted as $\{\mathbf{X}^v\in \mathbb{R}^{N\times D^v}  \}_{v=1}^V$, where $N$ and $V$ are the number of samples and views, respectively. $D^v$ represents the feature dimension of $v$-th view. A mask matrix $\mathbf{M} \in \mathbb{R}^{N\times V} $ restores the information that samples are observed or missing in specific views as,
\begin{equation}
	m_{iv} = 
	\begin{cases} 
		1 & \text{if } i\text{-th sample is observed in } v\text{-th view}, \\ 
		0 & \text{if } i\text{-th sample is absent in } v\text{-th view}, \\ 
	\end{cases}
\end{equation}
where $m_{iv}$ is the $iv$-th element of $\mathbf{M}$. Based on the mask $\mathbf{M}$, feature representations $\{\mathbf{X}^v\}_{v=1}^V$ can be divided into two parts, i.e. the observed parts $\{\bar{\mathbf{Z}}^v\in \mathbb{R}^{N_o^v\times D^v}  \}_{v=1}^V$ and view missing parts $\{\hat{\mathbf{Z}}^v\in \mathbb{R}^{N_m^v\times D^v}  \}_{v=1}^V$, where $N_o^v$ and $N_m^v$ denote the number of observed and absent samples in $v$-th view, respectively. In $\{\mathbf{X}^v\}_{v=1}^V$, the view missing parts $\{\hat{\mathbf{Z}}^v \}_{v=1}^V$ are unknown caused by some feature extraction failure cases. Note that when the mask $\mathbf{M}$ is an all-ones matrix, the data is referred to as complete multi-view data; otherwise, it is incomplete multi-view data.

To transfer the observed parts $\{\bar{\mathbf{Z}}^v \}_{v=1}^V$ and view missing parts $\{\hat{\mathbf{Z}}^v \}_{v=1}^V$ back into the original sample orders in $\{\mathbf{X}^v\}_{v=1}^V$, we formulate two kinds of permutation matrices, i.e, $\{\bar{\mathbf{T}}^v\in \mathbb{R}^{N_o^v\times N} \}_{v=1}^V$ and $\{\hat{\mathbf{T}}^v\in \mathbb{R}^{N_m^v\times N} \}_{v=1}^V$, in which their elements are defined as,
\begin{equation}
	\bar{t}_{ij}^v = 
	\begin{cases} 
		1 & \text{if } \bar{\mathbf{z}}_i^v \text{ is transferred to } \mathbf{x}_j^v, \\ 
		0 & \text{otherwise},
	\end{cases}
\end{equation}
\begin{equation}
	\hat{t}_{ij}^v = 
	\begin{cases} 
		1 & \text{if } \hat{\mathbf{z}}_i^v \text{ is transferred to } \mathbf{x}_j^v, \\ 
		0 & \text{otherwise},
	\end{cases}
\end{equation}
where $\bar{t}_{ij}^v$ and $\hat{t}_{ij}^v$ are the $ij$-th elements in $\bar{\mathbf{T}}^v$ and $\hat{\mathbf{T}}^v$, respectively. $\bar{\mathbf{z}}_i^v$ and $\hat{\mathbf{z}}_i^v$ represent the $i$-th samples of 
$\bar{\mathbf{Z}}^v$ and $\hat{\mathbf{Z}}^v$, respectively. $\mathbf{x}_j^v$ denotes the $j$-th sample of $\mathbf{X}^v$. Then, we can permute the observed parts $\{\bar{\mathbf{Z}}^v \}_{v=1}^V$ and view missing parts $\{\hat{\mathbf{Z}}^v \}_{v=1}^V$ back into the original feature orders as,
\begin{equation}
	\mathbf{X}^v =\bar{\mathbf{T}}^{v\top} \bar{\mathbf{Z}}^v   + \hat{\mathbf{T}}^{v\top} \hat{\mathbf{Z}}^v  
\end{equation}

\subsection{Overview}\label{sec:MMCF}

As shown in Fig.~\ref{fig:IMvC}, our proposed mask-informed deep contrastive incomplete multi-view clustering framework consists of three parts, including view-specific encoders, decoders, and mask-informed fusion.

\textbf{View-specific encoders}: The graph convolutional networks (GCNs) are widely used in many data mining tasks, aggregating the neighbor information to learn features~\cite{xia2021graph}. In our method, we first construct $k$-nearest neighbor graphs $\{\mathbf{A}^v \in \mathbb{R}^{N_o^v\times N_o^v}\}_{v=1}^V$ from the observed parts $\{\bar{\mathbf{Z}}\}_{v=1}^V$ by the adaptive neighbor learning method~\cite{nie2014clustering}. The view-specific encoders $\{f^v(\bar{\mathbf{Z}}^v, \mathbf{A}^v; \Theta^v)\}_{v=1}^V$, with $\Theta^v$ denoting the parameters, are implemented by several GCN layers to merge the neighbor information of each sample to learn view-specific latent representations. One of the GCN layers is formulated as follows,
\begin{equation}
	\mathbf{H}^{v} = \sigma({\mathbf{D}^{v}}^{-\frac{1}{2}}\bar{\mathbf{A}}^v{\mathbf{D}^{v}}^{-\frac{1}{2}}\mathbf{H}^{v}\mathbf{W}^{v}),
\end{equation}
where $\mathbf{H}^{v} \in \mathbb{R}^{N \times L} $ is the $v$-th view latent representation. $L$ denotes the feature dimension. $\bar{\mathbf{A}}^v = \mathbf{A}^v + \mathbf{I}$, with $\mathbf{I} \in \mathbb{R}^{N_o^v\times N_o^v}$ denoting an identity matrix. $\mathbf{D}^v$ is a degree matrix of $\mathbf{A}^v$.

\textbf{Mask-informed fusion}: After the view-specific feature extraction, the obtained multiple features are permuted as the same orders as the original features as by permutation matrices $\{\bar{\mathbf{T}}^v\}_{v=1}^V$ as,
\begin{equation}
	\bar{\mathbf{H}}^{v} = \bar{\mathbf{T}}^{v\top} \mathbf{H}^{v}.
\end{equation}

The mask $\mathbf{M}$ indicates whether there exist samples in specific views. Based on this knowledge, we exploit mask $\mathbf{M}$ to filter the missing value and remove their contributions during the multi-view feature aggregation procedure as follows,
\begin{equation}
	\mathbf{f}_i = \frac{\sum_{v=1}^v m_{iv} \bar{\mathbf{h}}^v_i}{\sum_{v=1}^v m_{iv}},
\end{equation}
where $\mathbf{f}_i$ is the $i$-th samples of view-common features $\mathbf{F}$. In such an above manner, incomplete information is automatically filtered to avoid adverse impacts on the view-common features. To further boost the representation capability of $\mathbf{F}$, we introduced a prior knowledge-assisted contrastive learning manner that is detailed depicted in Section~\ref{sec:PKCL}.

\textbf{View-specific decoders}: The view-specific decoders are introduced to reconstruct the observed parts in the original IMvC data. To this end, we first obtain the observed parts of each view in the view-common feature $\mathbf{F}$ by permutation matrices $\{\bar{\mathbf{T}}^v\}_{v=1}^V$ as,
\begin{equation}
	\bar{\mathbf{F}}^v = \bar{\mathbf{T}}^{v} \mathbf{F},
\end{equation}
where $\bar{\mathbf{F}}^v \in \mathbb{R}^{N_o \times L}$ is the obtained observed parts of each view in the view-common feature $\mathbf{F}$. Afterward, latent features pass to the view-specific decoders $\{g^v(\bar{\mathbf{F}}^v; \Phi^v)\}_{v=1}^V$ to reconstruct the view-specific features, which is optimized by minimizing the reconstruction loss between the original and reconstructed features as,
\begin{equation}
	\ell_{rec} = \frac{1}{N}\sum_{v=1}^V \sum_{i=1}^N \Vert \bar{\mathbf{z}}_i^v - g^v(\bar{\mathbf{f}}_i^v; \Phi^v) \Vert_2^2.
\end{equation}

\subsection{Prior Knowledge-assisted Contrastive Learning}\label{sec:PKCL}
The similarities among samples defined from the original features are critical prior knowledge for feature representation learning~\cite{he2003locality}. In this section, we inject the prior knowledge into the view-common feature representation with useful self-supervised contrastive learning to further boost its representation capability. To this end, we formulate three key components, i.e., the clustering projection layer, mask-informed graph fusion, and contrastive loss.

\textbf{Clustering projection layer:}
In this part, we first transfer the view-common representation into the cluster space with a single linear layer. Then, the QR decomposition~\cite{gander1980algorithms} is introduced to guarantee the orthogonal property of the clustering indicator matrix $\mathbf{Y} \in \mathbb{R}^{N \times C}$, where $C$ is the number of clusters. And the clustering distributions of the incomplete multi-view data can be revealed in $\mathbf{Y}$.

\textbf{Mask-informed graph fusion:}
To flexibly merge multiple similarity graphs constructed from IMvC data, view-specific graphs $\{\mathbf{A}^v\}_{v=1}^V$ are first transferred based on permutation matrices $\{\bar{\mathbf{T}}^v\}_{v=1}^V$ as,
\begin{equation}
	\bar{\mathbf{A}}^v = \bar{\mathbf{T}}^{v\top} \mathbf{A}^v \bar{\mathbf{T}}^v,
\end{equation}
where $\bar{\mathbf{A}}^v \in \mathbb{R}^{N\times N}$ is the transferred view-specific graph in $v$-view. Due to the incompleteness of data, $\{\bar{\mathbf{A}}^v\}_{v=1}^V$ only provides the sample similarities among the observed samples in each view, lacking the similarities among all samples. Thus, we construct a view-common similarity graph $\mathbf{S} \in \mathbb{R}^{N\times N}$ from view-common feature representation $\mathbf{F}$ to measure the sample correlations among all points. Simply averaging the matrices to fuse the sample similarities from multiple graphs will bring in the noise information caused by the incomplete values. Thus, we utilize the mask $\mathbf{M}$ to filter entries of missing values to achieve more effective fusion. To this end, the graph mask matrices $\{\mathbf{R}^v \in \mathbb{R}^{N \times N}\}_{v=1}^V$ is introduced based on the mask matrix $\mathbf{M}$ to indicate whether there exits sample correlations in $\{\bar{\mathbf{A}}^v\}_{v=1}^V$, computed as,
\begin{equation}
	r_{ij}^v = m_{iv} m_{jv},
\end{equation}
where $r_{ij}^v$ is the $ij$-th element of $\mathbf{R}^v$. Finally, the comprehensive similarity information among multiple graphs can be merged as,
\begin{equation}
	\hat{a}_{ij} = \frac{\sum_{v=1}^V r_{ij}^v \bar{a}_{ij}^v + s_{ij}}{\sum_{v=1}^V r_{ij}^v + 1},
\end{equation}
where $\hat{a}_{ij}$ is the $ij$-th element of graph $\hat{\mathbf{A}} \in \mathbb{R}^{N\times N}$.

\textbf{Contrastive loss:}
The contrastive learning approaches push the samples far away from the negative anchors while pulling in the samples with positive anchors~\cite{chen2020simple, wang2021understanding}. A straightforward way of constructing the positive and negative pairs is to leverage the sample connections in the graph $\hat{\mathbf{A}}$. Assuming that the neighbor points of each sample can be regarded as their positive anchors, and the points in the neighborhood of other samples can be treated as the negative anchors, we can construct the following contrastive loss to constrain the clustering distribution $\mathbf{Y}$ as formulation of InfoNCE~\cite{yeh2022decoupled} as~\footnote{The positive part is removed from denominator to alleviate the negative-positive-coupling issue.},
\begin{equation}\label{eq:cl_loos}
	\zeta_{dcl} \  =  - \frac{1}{N}\sum_{i=1}^N \text{log} \frac{\sum\limits_{j\in N^+} e^\frac{\text{sim}(\mathbf{y}_i, \mathbf{y}_j)}{\tau} }{\sum\limits_{r\in N^-} e^\frac{\text{sim}(\mathbf{y}_i, \mathbf{y}_r)}{\tau}},
\end{equation}
where $\text{sim}(\mathbf{y}_i, \mathbf{y}_j)$ denotes the cosine similarity computed as $\mathbf{y}_i\mathbf{y}_j^\top / \Vert\mathbf{y}_i \Vert \Vert\mathbf{y}_j \Vert$. $\tau$ is a temperature parameter. $N^+$ and $N^-$ respectively denote the positive and negative sets defined by sample connection relationships of the graph $\hat{\mathbf{A}}$.

Similarity graph $\hat{\mathbf{A}}$ measures similarities of samples as their connection possibilities in the range of $[0, 1]$ as defined in~\cite{nie2014clustering}. Simply treating the two samples with connection possibility $\hat{a}_{ij}$ larger than 0 as the positive pair is not suitable due to there is also $1 - \hat{a}_{ij}$ possibility that can be regarded as negative pair. Therefore, we can further re-weight the Eq.~\eqref{eq:cl_loos} from a binary case into continual version as,
\begin{equation}\label{eq:vcrn_loss}
	\zeta_{wcl} \  =  - \frac{1}{N} \sum_{i=1}^N \text{log} \frac{\sum\limits_{j=1}^N \hat{a}_{ij} e^\frac{\text{sim}(\mathbf{y}_i, \mathbf{y}_j)}{\tau}} {\sum\limits_{r=1}^N (1-\hat{a}_{ir}) e^\frac{\text{sim}(\mathbf{y}_i, \mathbf{y}_r)}{\tau}}.
\end{equation}

As shown in Eq.~\eqref{eq:vcrn_loss}, the positive and negative pairs can be adaptively determined by the similarity graph $\hat{\mathbf{A}}$. In addition, the samples with similarities equal to zeros are automatically treated as negative pairs. The samples with similarities larger than zeros ensure the possibility of being grouped into either positive or negative pairs that are re-weighted with their corresponding possibilities. Under this contrastive loss, the view-common feature representation preserves the neighbor information among different samples and reveals more compact and clear structures for clustering.

\subsection{Total Loss Function}\label{sec:trga}
Finally, the total loss function of the proposed model is formulated as,
\begin{equation}
	\mathcal{L} = \ell_{rec} + \lambda \zeta_{wcl}
\end{equation}
where $\lambda$ is a hyper-parameter to balance reconstruction loss $\ell_{rec}$ and re-weighted contrastive loss $\zeta_{wcl}$.

\section{Experiments}

\subsection{Experimental Settings}

\textbf{Datasets}:
In our experiments, we evaluate our proposed method on four widely used multi-view datasets, including \textbf{MSRCV}~\footnote{https://mldta.com/dataset/msrc-v1/}, \textbf{CUB}~\footnote{https://www.vision.caltech.edu/datasets/cub\_200\_2011/}, 
\textbf{OutdoorScene}~\cite{hu2020multi},
and \textbf{nuswide}~\cite{zhen2019deep}. The details descriptions of the datasets are given in follows:

\textbf{MSRCV}: This dataset contains 210 images corresponding to seven different classes. Each image is characterized by three views: 256-dimensional LBP, 512-dimensional GIST, and 210-dimensional SIFT features.

\textbf{CUB}: This dataset consists of 600 images of different bird species, each accompanied by text descriptions, spanning 10 categories. Each sample is represented by 4096-dimensional deep image features and 300-dimensional text features as doing in~\cite{zhen2019deep}.

\textbf{OutdoorScene}: This dataset includes 2688 images of outdoor scenes captured from eight different scene groups. Each image is described using four views: 512-dimensional GIST features, 432-dimensional HOG features, 256-dimensional LBP features, and 48-dimensional Gabor features.

\textbf{nuswide}: This dataset includes 9000 images along with corresponding tags from 10 categories. Each sample is represented by 4096-dimensional deep image features and 300-dimensional text features as doing in~\cite{zhen2019deep}. We randomly selected 3000 samples in our experiments.

\begin{table*}[t]
	\centering
	\caption{The clustering performance measured by ACC, NMI, ARI, and Fscore of all compared methods on multi-view datasets. The highest and the second highest values under each metric are {\color[HTML]{EA6B66}\textbf{bolded}} and {\color[HTML]{67AB9F}\ul{ underlined}}, respectively.}
	\label{tab:mvc_results}
	\resizebox{\textwidth}{!}{%
		\begin{tabular}{@{}c|cccc|cccc|cccc|cccc@{}}
			\toprule[1.5pt]
			Datasets & \multicolumn{4}{c|}{CUB}                                                                                                                                  & \multicolumn{4}{c|}{MSRCV}                                                                                                                                & \multicolumn{4}{c|}{OutdoorScene}                                                                                                                         & \multicolumn{4}{c}{nuswide}                                                                                                                               \\ \midrule
			Methods  & ACC                                  & NMI                                  & ARI                                  & Fscore                               & ACC                                  & NMI                                  & ARI                                  & Fscore                               & ACC                                  & NMI                                  & ARI                                  & Fscore                               & ACC                                  & NMI                                  & ARI                                  & Fscore                               \\ \midrule
			DCCA     & 60.09                                & 55.87                                & 40.79                                & 48.84                                & 68.90                                & 62.09                                & 50.06                                & 61.84                                & 48.16                                & 51.58                                & 36.42                                & 49.76                                & 48.16                                & 40.03                                & 22.89                                & 41.40                                \\
			AWP      & 81.17                                & 75.20                                & 65.52                                & 69.00                                & 67.62                                & 62.16                                & 53.58                                & 60.23                                & 60.83                                & 45.51                                & 37.37                                & 45.98                                & 55.93                                & 47.05                                & 36.38                                & 44.37                                \\
			GMC      & 79.50                                & 78.95                                & 66.48                                & 70.03                                & 73.81                                & 74.32                                & 64.03                                & 69.41                                & 33.56                                & 42.33                                & 18.78                                & 35.02                                & 22.83                                & 19.60                                & 1.72                                 & 19.68                                \\
			OPMC     & 71.50                                & 75.67                                & 60.92                                & 64.95                                & 82.38                                & 73.38                                & 66.96                                & 71.63                                & 63.17                                & 51.57                                & 42.22                                & 49.62                                & 58.60                                & 44.87                                & 37.35                                & 43.82                                \\
			MFLVC    & 67.00                                & 62.91                                & 49.24                                & 55.59                                & 63.33                                & 55.42                                & 45.23                                & 57.13                                & 64.62                                & 55.15                                & 42.90                                & 51.11                                & 32.10                                & 20.75                                & 16.09                                & 24.40                                \\
			DealMVC  & 61.00                                & 66.10                                & 51.27                                & 58.04                                & 44.29                                & 31.91                                & 19.54                                & 37.77                                & {\color[HTML]{67AB9F}\ul{75.26}}     & \color[HTML]{EA6B66}\textbf{{62.73}} & \color[HTML]{EA6B66}\textbf{{54.21}} & \color[HTML]{EA6B66}\textbf{{61.83}} & 51.93                                & 37.88                                & 32.75                                & 38.81                                \\
			CVCL     & 79.33                                & 71.03                                & 61.54                                & 66.17                                & 72.00                                & 58.60                                & 50.23                                & 59.08                                & 71.65                                & 60.23                                & 50.34                                & 57.98                                & 58.63                                & {\color[HTML]{67AB9F}\ul{48.30}}     & {\color[HTML]{67AB9F}\ul{39.92}}     & {\color[HTML]{67AB9F}\ul{47.41}}     \\
			EEOMVC   & 65.00                                & 66.49                                & 51.32                                & 56.43                                & 69.05                                & 54.64                                & 45.63                                & 53.30                                & 57.70                                & 40.35                                & 29.64                                & 39.02                                & {\color[HTML]{67AB9F}\ul{61.27}}     & 45.02                                & 39.80                                & 46.09                                \\
			SCMVC    & 72.83                                & 68.71                                & 55.82                                & 61.66                                & 71.90                                & 63.02                                & 56.26                                & 63.36                                & 72.10                                & 60.57                                & 52.56                                & 59.89                                & 53.97                                & 40.63                                & 34.91                                & 41.37                                \\
			SCM      & \color[HTML]{EA6B66}\textbf{{84.83}} & {\color[HTML]{67AB9F}\ul{79.11}}     & {\color[HTML]{67AB9F}\ul{71.23}}     & {\color[HTML]{67AB9F}\ul{75.17}}     & {\color[HTML]{67AB9F}\ul{89.05}}     & {\color[HTML]{67AB9F}\ul{80.95}}     & {\color[HTML]{67AB9F}\ul{76.66}}     & {\color[HTML]{67AB9F}\ul{81.01}}     & 60.45                                & 51.95                                & 41.59                                & 49.93                                & 57.63                                & 44.17                                & 38.51                                & 44.56                                \\ \midrule
			\rowcolor[HTML]{EFEFEF} 
			Ours     & {\color[HTML]{67AB9F}\ul{81.98}}     & \color[HTML]{EA6B66}\textbf{{85.33}} & \color[HTML]{EA6B66}\textbf{{76.23}} & \color[HTML]{EA6B66}\textbf{{81.97}} & \color[HTML]{EA6B66}\textbf{{90.76}} & \color[HTML]{EA6B66}\textbf{{81.84}} & \color[HTML]{EA6B66}\textbf{{79.28}} & \color[HTML]{EA6B66}\textbf{{83.37}} & \color[HTML]{EA6B66}\textbf{{75.53}} & {\color[HTML]{67AB9F}\ul{61.19}}     & {\color[HTML]{67AB9F}\ul{54.13}}     & {\color[HTML]{67AB9F}\ul{61.54}}     & \color[HTML]{EA6B66}\textbf{{64.34}} & \color[HTML]{EA6B66}\textbf{{49.89}} & \color[HTML]{EA6B66}\textbf{{41.33}} & \color[HTML]{EA6B66}\textbf{{50.32}} \\ \bottomrule[1.5pt]
		\end{tabular}%
	}
\end{table*}

\begin{table*}[!htbp]
	\centering
	\caption{The clustering performance measured by ACC, NMI, ARI, and Fscore of all compared methods on multi-view datasets with different missing rates $\eta$. The highest and the second highest values under each metric are {\color[HTML]{EA6B66}\textbf{bolded}} and {\color[HTML]{67AB9F}\ul{ underlined}}, respectively.}
	\label{tab:imvc_results}
	\resizebox{\textwidth}{!}{%
		\begin{tabular}{@{}c|c|cccc|cccc|cccc|cccc@{}}
			\toprule[1.5pt]
			& Datasets                     & \multicolumn{4}{c|}{CUB}                                                                                                                                                                                                                          & \multicolumn{4}{c|}{MSRCV}                                                                                                                                                                                                                                & \multicolumn{4}{c|}{OutdoorScene}                                                                                                                                                                                                                         & \multicolumn{4}{c}{nuswide}                                                                                                                                                                                                                               \\ \cmidrule(l){2-18} 
			\multirow{-2}{*}{$\eta$} & Methods                      & ACC                                                      & NMI                                                      & ARI                                                          & Fscore                                                       & ACC                                                          & NMI                                                          & ARI                                                          & Fscore                                                       & ACC                                                          & NMI                                                          & ARI                                                          & Fscore                                                       & ACC                                                          & NMI                                                          & ARI                                                          & Fscore                                                       \\ \midrule
			& EFAE                         & 64.94                                                    & 64.84                                                    & 51.83                                                        & 59.76                                                        & 70.35                                                        & 59.67                                                        & 51.94                                                        & 60.55                                                        & 63.15                                                        & 51.26                                                        & 42.65                                                        & 51.83                                                        & 51.69                                                        & 37.06                                                        & 29.97                                                        & 38.80                                                        \\
			& MFAE                         & 62.35                                                    & 64.65                                                    & 51.14                                                        & 58.52                                                        & 66.11                                                        & 58.03                                                        & 49.16                                                        & 59.30                                                        & 62.31                                                        & 50.74                                                        & 41.33                                                        & 50.53                                                        & 53.61                                                        & 41.14                                                        & 33.41                                                        & {\color[HTML]{67AB9F}\ul{42.07}}                             \\
			& DCCA                         & 33.01                                                    & 28.50                                                    & 11.02                                                        & 29.80                                                        & 46.69                                                        & 38.94                                                        & 21.14                                                        & 44.58                                                        & 50.01                                                        & 49.53                                                        & 34.41                                                        & 47.58                                                        & 34.51                                                        & 24.30                                                        & 10.78                                                        & 31.46                                                        \\
			& DSIMVC                       & 56.07                                                    & 56.31                                                    & 38.85                                                        & 45.66                                                        & 47.62                                                        & 38.88                                                        & 28.21                                                        & 39.34                                                        & 54.99                                                        & 50.76                                                        & 35.87                                                        & 44.23                                                        & 33.57                                                        & 25.19                                                        & 16.53                                                        & 24.86                                                        \\
			& DSIMVC++                     & 59.70                                                    & 57.25                                                    & 41.07                                                        & 47.52                                                        & 44.76                                                        & 37.65                                                        & 24.85                                                        & 36.85                                                        & 54.89                                                        & 49.23                                                        & 35.27                                                        & 43.63                                                        & 41.32                                                        & 35.12                                                        & 24.15                                                        & 31.58                                                        \\
			& CPSPAN                       & 63.37                                                    & 62.80                                                    & 48.18                                                        & 60.86                                                        & 73.43                                                        & 66.40                                                        & 56.24                                                        & 70.87                                                        & 57.96                                                        & 49.55                                                        & 37.15                                                        & {\color[HTML]{67AB9F}\ul{57.83}}                             & 39.47                                                        & 28.74                                                        & 19.00                                                        & 39.25                                                        \\
			& GIGA                         & 73.77                                                    & \color[HTML]{EA6B66}\textbf{{77.54}}                     & 63.42                                                        & {\color[HTML]{67AB9F}\ul{71.96}}                             & 74.48                                                        & {\color[HTML]{67AB9F}\ul{76.69}}                             & {\color[HTML]{67AB9F}\ul{62.52}}                             & {\color[HTML]{67AB9F}\ul{76.14}}                             & 46.57                                                        & 54.58                                                        & 35.13                                                        & 52.72                                                        & 30.61                                                        & 30.54                                                        & 7.18                                                         & 41.01                                                        \\
			& MVCAN                        & 60.34                                                    & 62.40                                                    & 49.09                                                        & 56.36                                                        & 65.75                                                        & 55.97                                                        & 46.67                                                        & 57.46                                                        & {\color[HTML]{67AB9F}\ul{65.50}}                             & {\color[HTML]{67AB9F}\ul{56.62}}                             & {\color[HTML]{67AB9F}\ul{46.39}}                             & 54.68                                                        & 49.05                                                        & 36.51                                                        & 28.85                                                        & 37.44                                                        \\
			& SCM                          & \color[HTML]{EA6B66}\textbf{{79.37}}                     & 73.75                                                    & {\color[HTML]{67AB9F}\ul{64.10}}                             & 69.01                                                        & {\color[HTML]{67AB9F}\ul{76.95}}                             & 67.85                                                        & 60.35                                                        & 67.63                                                        & 59.73                                                        & 48.46                                                        & 37.72                                                        & 47.26                                                        & {\color[HTML]{67AB9F}\ul{56.43}}                             & {\color[HTML]{67AB9F}\ul{41.29}}                             & {\color[HTML]{67AB9F}\ul{34.69}}                             & 41.80                                                        \\ \cmidrule(l){2-18} 
			\multirow{-10}{*}{0.10}  & \cellcolor[HTML]{EFEFEF}Ours & \cellcolor[HTML]{EFEFEF}{\color[HTML]{67AB9F}\ul{78.95}} & \cellcolor[HTML]{EFEFEF}{\color[HTML]{67AB9F}\ul{77.31}} & \cellcolor[HTML]{EFEFEF}\color[HTML]{EA6B66}\textbf{{65.62}} & \cellcolor[HTML]{EFEFEF}\color[HTML]{EA6B66}\textbf{{72.37}} & \cellcolor[HTML]{EFEFEF}\color[HTML]{EA6B66}\textbf{{89.69}} & \cellcolor[HTML]{EFEFEF}\color[HTML]{EA6B66}\textbf{{80.01}} & \cellcolor[HTML]{EFEFEF}\color[HTML]{EA6B66}\textbf{{77.20}} & \cellcolor[HTML]{EFEFEF}\color[HTML]{EA6B66}\textbf{{81.63}} & \cellcolor[HTML]{EFEFEF}\color[HTML]{EA6B66}\textbf{{74.20}} & \cellcolor[HTML]{EFEFEF}\color[HTML]{EA6B66}\textbf{{61.62}} & \cellcolor[HTML]{EFEFEF}\color[HTML]{EA6B66}\textbf{{51.96}} & \cellcolor[HTML]{EFEFEF}\color[HTML]{EA6B66}\textbf{{61.04}} & \cellcolor[HTML]{EFEFEF}\color[HTML]{EA6B66}\textbf{{60.67}} & \cellcolor[HTML]{EFEFEF}\color[HTML]{EA6B66}\textbf{{46.15}} & \cellcolor[HTML]{EFEFEF}\color[HTML]{EA6B66}\textbf{{39.64}} & \cellcolor[HTML]{EFEFEF}\color[HTML]{EA6B66}\textbf{{46.74}} \\ \bottomrule

			\addlinespace[0.3em]
			
			\toprule
			& EFAE                         & 46.39                                                        & 49.34                                                    & 31.34                                                    & 41.60                                                    & 58.98                                                        & 47.63                                                    & 38.47                                                        & 48.98                                                        & 55.01                                                        & 43.38                                                        & 35.83                                                        & 45.27                                                        & 41.30                                                        & 29.18                                                        & 19.68                                                        & 29.91                                                        \\
			& MFAE                         & 46.24                                                        & 49.07                                                    & 30.93                                                    & 40.81                                                    & 55.01                                                        & 45.29                                                    & 35.56                                                        & 46.83                                                        & 48.79                                                        & 39.34                                                        & 32.23                                                        & 41.24                                                        & 41.45                                                        & 31.16                                                        & 20.84                                                        & 30.93                                                        \\
			& DCCA                         & 19.31                                                        & 12.90                                                    & 1.83                                                     & 24.31                                                    & 22.10                                                        & 12.65                                                    & 1.51                                                         & 31.81                                                        & 49.99                                                        & 46.92                                                        & 32.27                                                        & 45.53                                                        & 20.31                                                        & 7.42                                                         & 1.79                                                         & 20.15                                                        \\
			& DSIMVC                       & 53.13                                                        & 51.93                                                    & 34.17                                                    & 41.58                                                    & 44.10                                                        & 38.57                                                    & 26.56                                                        & 38.50                                                        & 52.89                                                        & 48.83                                                        & 33.50                                                        & 42.18                                                        & 26.55                                                        & 17.37                                                        & 11.16                                                        & 21.16                                                        \\
			& DSIMVC++                     & 61.77                                                        & 58.04                                                    & 42.58                                                    & 48.88                                                    & 52.10                                                        & 43.77                                                    & 32.54                                                        & 42.97                                                        & 53.92                                                        & 48.83                                                        & 34.42                                                        & 42.78                                                        & 45.41                                                        & {\color[HTML]{67AB9F}\ul{37.04}}                             & {\color[HTML]{67AB9F}\ul{26.84}}                             & 34.22                                                        \\
			& CPSPAN                       & 60.37                                                        & 59.56                                                    & 43.70                                                    & 59.62                                                    & 71.24                                                        & 65.47                                                    & 54.22                                                        & 68.34                                                        & 56.05                                                        & 48.13                                                        & 35.32                                                        & {\color[HTML]{67AB9F}\ul{56.02}}                             & 39.67                                                        & 28.63                                                        & 19.44                                                        & {\color[HTML]{67AB9F}\ul{39.25}}                             \\
			& GIGA                         & {\color[HTML]{67AB9F}\ul{71.47}}                             & \color[HTML]{EA6B66}\textbf{{74.48}}                     & \color[HTML]{EA6B66}\textbf{{60.16}}                     & \color[HTML]{EA6B66}\textbf{{69.15}}                     & 70.76                                                        & \color[HTML]{EA6B66}\textbf{{73.16}}                     & 56.96                                                        & {\color[HTML]{67AB9F}\ul{72.55}}                             & 43.65                                                        & {\color[HTML]{67AB9F}\ul{49.82}}                             & 28.52                                                        & 51.73                                                        & 28.57                                                        & 26.90                                                        & 6.15                                                         & 38.33                                                        \\
			& MVCAN                        & 49.23                                                        & 49.77                                                    & 33.66                                                    & 42.41                                                    & 47.17                                                        & 36.39                                                    & 26.68                                                        & 39.37                                                        & 51.67                                                        & 45.57                                                        & 35.23                                                        & 44.10                                                        & 37.12                                                        & 26.24                                                        & 16.90                                                        & 27.65                                                        \\
			& SCM                          & 68.63                                                        & 63.79                                                    & 50.93                                                    & 58.65                                                    & {\color[HTML]{67AB9F}\ul{80.10}}                             & 66.44                                                    & {\color[HTML]{67AB9F}\ul{60.54}}                             & 67.77                                                        & {\color[HTML]{67AB9F}\ul{59.51}}                             & 47.66                                                        & {\color[HTML]{67AB9F}\ul{36.72}}                             & 46.29                                                        & {\color[HTML]{67AB9F}\ul{47.21}}                             & 33.45                                                        & 24.57                                                        & 33.75                                                        \\ \cmidrule(l){2-18} 
			\multirow{-10}{*}{0.30}  & \cellcolor[HTML]{EFEFEF}Ours & \cellcolor[HTML]{EFEFEF}\color[HTML]{EA6B66}\textbf{{73.91}} & \cellcolor[HTML]{EFEFEF}{\color[HTML]{67AB9F}\ul{71.60}} & \cellcolor[HTML]{EFEFEF}{\color[HTML]{67AB9F}\ul{57.89}} & \cellcolor[HTML]{EFEFEF}{\color[HTML]{67AB9F}\ul{66.94}} & \cellcolor[HTML]{EFEFEF}\color[HTML]{EA6B66}\textbf{{82.62}} & \cellcolor[HTML]{EFEFEF}{\color[HTML]{67AB9F}\ul{72.31}} & \cellcolor[HTML]{EFEFEF}\color[HTML]{EA6B66}\textbf{{65.80}} & \cellcolor[HTML]{EFEFEF}\color[HTML]{EA6B66}\textbf{{73.09}} & \cellcolor[HTML]{EFEFEF}\color[HTML]{EA6B66}\textbf{{67.58}} & \cellcolor[HTML]{EFEFEF}\color[HTML]{EA6B66}\textbf{{56.93}} & \cellcolor[HTML]{EFEFEF}\color[HTML]{EA6B66}\textbf{{46.71}} & \cellcolor[HTML]{EFEFEF}\color[HTML]{EA6B66}\textbf{{57.22}} & \cellcolor[HTML]{EFEFEF}\color[HTML]{EA6B66}\textbf{{53.33}} & \cellcolor[HTML]{EFEFEF}\color[HTML]{EA6B66}\textbf{{40.48}} & \cellcolor[HTML]{EFEFEF}\color[HTML]{EA6B66}\textbf{{28.82}} & \cellcolor[HTML]{EFEFEF}\color[HTML]{EA6B66}\textbf{{40.21}} \\ \bottomrule
			
			\addlinespace[0.3em]
			
			\toprule
			& EFAE                         & 37.11                                                    & 40.26                                                    & 17.33                                                    & 32.59                                                    & 48.29                                                        & 38.13                                                    & 26.97                                                        & 40.02                                                    & 48.69                                                        & 37.15                                                    & 28.48                                                        & 38.64                                & 28.95                                                        & 21.22                                                    & 9.41                                 & 23.05                                \\
			& MFAE                         & 37.80                                                    & 41.23                                                    & 18.41                                                    & 32.78                                                    & 45.95                                                        & 33.73                                                    & 22.31                                                        & 36.63                                                    & 40.14                                                        & 29.92                                                    & 22.45                                                        & 32.39                                & 31.17                                                        & 24.91                                                    & 11.45                                & 25.24                                \\
			& DCCA                         & 14.01                                                    & 5.30                                                     & 0.29                                                     & 20.23                                                    & 19.08                                                        & 7.44                                                     & 0.71                                                         & 28.58                                                    & 47.49                                                        & 42.76                                                    & 29.95                                                        & 42.57                                & 14.24                                                        & 2.17                                                     & 0.18                                 & 18.11                                \\
			& DSIMVC                       & 52.82                                                    & 51.70                                                    & 34.07                                                    & 41.55                                                    & 43.52                                                        & 35.45                                                    & 23.34                                                        & 36.06                                                    & 52.98                                                        & 47.91                                                    & 32.92                                                        & 41.81                                & 27.78                                                        & 17.24                                                    & 11.54                                & 21.24                                \\
			& DSIMVC++                     & 60.30                                                    & 56.93                                                    & 40.83                                                    & 47.29                                                    & 47.33                                                        & 38.09                                                    & 26.60                                                        & 38.28                                                    & 52.37                                                        & 47.10                                                    & 32.94                                                        & 41.45                                & {\color[HTML]{67AB9F}\ul{42.70}}                             & \color[HTML]{EA6B66}\textbf{{34.24}}                     & \color[HTML]{EA6B66}\textbf{{24.45}} & 31.68                                \\
			& CPSPAN                       & 61.27                                                    & 61.52                                                    & 45.93                                                    & 59.29                                                    & 70.09                                                        & 62.06                                                    & 50.84                                                        & 67.23                                                    & 56.98                                                        & 47.45                                                    & 35.47                                                        & \color[HTML]{EA6B66}\textbf{{56.83}} & 42.64                                                        & 30.34                                                    & {\color[HTML]{67AB9F}\ul{21.81}}     & \color[HTML]{EA6B66}\textbf{{42.53}} \\
			& GIGA                         & \color[HTML]{EA6B66}\textbf{{69.30}}                     & \color[HTML]{EA6B66}\textbf{{70.35}}                     & \color[HTML]{EA6B66}\textbf{{54.35}}                     & \color[HTML]{EA6B66}\textbf{{64.85}}                     & 72.19                                                        & \color[HTML]{EA6B66}\textbf{{70.13}}                     & {\color[HTML]{67AB9F}\ul{54.00}}                             & \color[HTML]{EA6B66}\textbf{{70.11}}                     & 48.57                                                        & \color[HTML]{EA6B66}\textbf{{54.97}}                     & 34.06                                                        & {\color[HTML]{67AB9F}\ul{54.36}}     & 33.51                                                        & 30.91                                                    & 7.23                                 & {\color[HTML]{67AB9F}\ul{41.37}}     \\
			& MVCAN                        & 37.69                                                    & 42.19                                                    & 21.38                                                    & 31.93                                                    & 38.13                                                        & 25.14                                                    & 15.08                                                        & 30.25                                                    & 42.61                                                        & 35.92                                                    & 26.45                                                        & 36.10                                & 30.59                                                        & 21.02                                                    & 10.17                                & 23.83                                \\
			& SCM                          & 61.83                                                    & 59.83                                                    & 45.41                                                    & 52.21                                                    & {\color[HTML]{67AB9F}\ul{73.71}}                             & 58.33                                                    & 50.30                                                        & 59.84                                                    & {\color[HTML]{67AB9F}\ul{59.19}}                             & 48.32                                                    & {\color[HTML]{67AB9F}\ul{38.49}}                             & 47.36                                & 41.76                                                        & 28.57                                                    & 15.88                                & 29.80                                \\ \cmidrule(l){2-18} 
			\multirow{-10}{*}{0.50}  & \cellcolor[HTML]{EFEFEF}Ours & \cellcolor[HTML]{EFEFEF}{\color[HTML]{67AB9F}\ul{68.46}} & \cellcolor[HTML]{EFEFEF}{\color[HTML]{67AB9F}\ul{68.39}} & \cellcolor[HTML]{EFEFEF}{\color[HTML]{67AB9F}\ul{52.74}} & \cellcolor[HTML]{EFEFEF}{\color[HTML]{67AB9F}\ul{63.05}} & \cellcolor[HTML]{EFEFEF}\color[HTML]{EA6B66}\textbf{{80.25}} & \cellcolor[HTML]{EFEFEF}{\color[HTML]{67AB9F}\ul{69.42}} & \cellcolor[HTML]{EFEFEF}\color[HTML]{EA6B66}\textbf{{60.73}} & \cellcolor[HTML]{EFEFEF}{\color[HTML]{67AB9F}\ul{70.08}} & \cellcolor[HTML]{EFEFEF}\color[HTML]{EA6B66}\textbf{{60.84}} & \cellcolor[HTML]{EFEFEF}{\color[HTML]{67AB9F}\ul{49.47}} & \cellcolor[HTML]{EFEFEF}\color[HTML]{EA6B66}\textbf{{38.51}} & \cellcolor[HTML]{EFEFEF}50.19        & \cellcolor[HTML]{EFEFEF}\color[HTML]{EA6B66}\textbf{{45.51}} & \cellcolor[HTML]{EFEFEF}{\color[HTML]{67AB9F}\ul{34.12}} & \cellcolor[HTML]{EFEFEF}19.13        & \cellcolor[HTML]{EFEFEF}34.96        \\ \bottomrule[1.5pt]
			
		\end{tabular}%
	}
\end{table*}

\begin{table*}[!htbp]
	\centering
	\caption{The ablation study of the proposed method in terms of ACC, NMI, ARI, and Fscore on four benchmark multi-view datasets.}
	\label{tab:imvc_ab}
	\resizebox{\textwidth}{!}{%
		\begin{tabular}{@{}lllllllllllllllll@{}}
			\toprule[1.5pt]
			\multicolumn{1}{c|}{Datasets} & \multicolumn{4}{c|}{CUB}                            & \multicolumn{4}{c|}{MSRCV}                          & \multicolumn{4}{c|}{OutdoorScene}                   & \multicolumn{4}{c}{nuswide}    \\ \midrule
			\multicolumn{1}{c|}{Methods}  & ACC   & NMI   & ARI   & \multicolumn{1}{c|}{Fscore} & ACC   & NMI   & ARI   & \multicolumn{1}{c|}{Fscore} & ACC   & NMI   & ARI   & \multicolumn{1}{c|}{Fscore} & ACC   & NMI   & ARI   & Fscore \\ \midrule
			\multicolumn{1}{c|}{Ours}     & 73.91 & 71.60 & 57.89 & \multicolumn{1}{c|}{66.94}  & 82.62 & 72.31 & 65.80 & \multicolumn{1}{c|}{73.09}  & 67.58 & 56.93 & 46.71 & \multicolumn{1}{c|}{57.22}  & 53.33 & 40.48 & 28.82 & 40.21  \\ \midrule
			\multicolumn{17}{l}{(a) Effectiveness   of Mask-informed Fusion Strategy}                                                                                                                                                        \\ \midrule
			\multicolumn{1}{c|}{wo-MFF}   & 65.34 & 64.58 & 47.57 & \multicolumn{1}{c|}{57.79}  & 79.36 & 72.28 & 62.89 & \multicolumn{1}{c|}{72.81}  & 63.74 & 53.27 & 44.09 & \multicolumn{1}{c|}{53.76}  & 51.52 & 40.19 & 28.35 & 39.78  \\
			\multicolumn{1}{c|}{wo-MGF}   & 70.76 & 69.16 & 53.69 & \multicolumn{1}{c|}{64.29}  & 80.97 & 71.28 & 64.45 & \multicolumn{1}{c|}{72.36}  & 66.45 & 55.88 & 47.72 & \multicolumn{1}{c|}{56.32}  & 51.98 & 39.40 & 27.05 & 39.26  \\ \midrule
			\multicolumn{17}{l}{(b) Effectiveness   of Prior Knowledge-assisted Contrastive Learning}                                                                                                                                        \\ \midrule
			\multicolumn{1}{c|}{wo-WCL}   & 40.77 & 50.38 & 32.07 & \multicolumn{1}{c|}{40.74}  & 60.88 & 50.15 & 40.21 & \multicolumn{1}{c|}{52.32}  & 57.64 & 48.04 & 38.84 & \multicolumn{1}{c|}{48.45}  & 38.63 & 31.78 & 20.41 & 33.51  \\
			\multicolumn{1}{c|}{wo-Rec}   & 74.29 & 71.72 & 57.65 & \multicolumn{1}{c|}{66.79}  & 75.57 & 69.26 & 58.35 & \multicolumn{1}{c|}{68.27}  & 67.97 & 57.10 & 47.11 & \multicolumn{1}{c|}{57.33}  & 52.96 & 41.37 & 24.62 & 41.59  \\
			\multicolumn{1}{c|}{w-CL}     & 41.40 & 51.28 & 33.24 & \multicolumn{1}{c|}{41.22}  & 66.18 & 54.67 & 45.04 & \multicolumn{1}{c|}{56.68}  & 58.16 & 49.08 & 39.91 & \multicolumn{1}{c|}{49.04}  & 38.36 & 30.03 & 19.20 & 31.90  \\
			\multicolumn{1}{c|}{w-DCL}    & 64.01 & 66.79 & 50.53 & \multicolumn{1}{c|}{58.63}  & 77.07 & 65.02 & 57.19 & \multicolumn{1}{c|}{65.65}  & 68.09 & 57.28 & 44.96 & \multicolumn{1}{c|}{54.92}  & 51.69 & 39.76 & 27.76 & 40.00  \\ \bottomrule[1.5pt]
		\end{tabular}%
	}
\end{table*}

\textbf{Evaluation metrics}:
Four widely used evaluation metrics, i.e., accuracy (ACC), normalized mutual information (NMI), adjusted Rand index (ARI), and Fscore are utilized to evaluate the clustering performance. And higher value of each metric denotes a better clustering result.

\subsection{Implementation details}
The detailed structures of the view-specific encoders and decoders are respectively $D$-196-128-64 and 64-128-196-$D$, where $D$ denotes the input feature dimension. The PyTorch~\cite{paszke2019pytorch} tool is utilized to implement the proposed method, in which the parameters are optimized with Adam optimizer~\cite{kingma2014adam} with 0.001 initial learning rate. The proposed model is trained for 1500 epochs on all datasets, and the aggregated view-common feature representation is used to capture the cluster structures with the K-means. The features in all datasets are scaled into the range $[0, 1]$ in our experiments. the trade-off parameter $\lambda$ in the proposed method is tuned in the range of $[10^{-3}, 10^{-2}, ...,10^2, 10^3]$. Temperature $\tau$ is set to 1. The number of neighbors in the formulated graphs is set to 15 for all datasets. All experiments are conducted on a single NVIDIA 2080Ti GPU with a Ubuntu 20.04 platform.

\subsection{Experimental Results}
To verify the effectiveness of the proposed method, we perform experiments on both view-complete and view-missing settings with various approaches. The detailed results and analysis are given in the following parts.

\textbf{Results on complete MvC datasets}:
In this part, we compared the proposed method on view-complete MvC datasets with ten different approaches, including DCCA~\cite{wang2015deep}, AWP~\cite{nie2018multiview}, GMC~\cite{wang2019gmc}, OPMC~\cite{liu2021one}, MFLVC~\cite{xu2022multi}, DealMVC~\cite{chen2023deep}, CVCL~\cite{chen2023deep}, EEOMVC~\cite{wang2023efficient}, SCMVC~\cite{wu2024self} and SCM~\cite{luo2024simple}. 
The parameters in all compared approaches are tuned with a grid search scheme as suggested in their papers to implement their best clustering performance. In Tab.~\ref{tab:mvc_results}, we present the clustering results in terms of four metrics on four MvC datasets of all methods.

From the Tab.~\ref{tab:mvc_results}, we obtain the following observations: 
Our method consistently performs better than other competitors on four datasets. For example, the proposed method outperforms the second performer, e.g. SCM method, with about 1.71, 0.98, 2.62, and 2.36 percentages in terms of ACC, NMI, ARI, and Fscore on the MSRCV dataset. Such advantages can be observed consistently on other datasets, which indicates the superiority of the proposed method. There are some methods, e.g. MFLVC, DealMVC, CVCL, and SCM, built upon contrastive learning. They usually treat the sample samples in different views as the positive pairs, while formulating the different samples in diverse views as the negative pairs. This type of contrastive learning manner inevitably ignores the structural information among multi-view datasets, resulting in sub-optimal clustering performance. Differently, the proposed method constructs the positive and negative pairs based on the similarities among different samples and further introduces a weighted contrastive loss to guide the model to learn more discriminative view-common feature representation for clustering. The deep MvC methods, e.g., Ours, SCM, and SCMVC perform better than traditional MvC approaches, e.g. AWP, GMC, and OPMC, demonstrating the effectiveness of deep feature representation. The above superior performance of the proposed method strongly verifies the effectiveness of the proposed method.

\textbf{Results on incomplete MvC datasets}:
We verify the superiority of the proposed method on incomplete MvC datasets with nine different approaches: 
EEAE, MFAE, DCCA~\cite{wang2015deep}, DSIMVC~\cite{tang2022deep}, DSIMVC++~\cite{Yan_2023_CVPR}, CPSPAN~\cite{jin2023deep}, GIGA~\cite{yang2024geometric}, MVCAN~\cite{xu2024investigating} and SCM~\cite{luo2024simple}. 
The parameters in all compared approaches are tuned with a grid search scheme as suggested in their papers to implement their best clustering performance. In our experiments, we follow the strategy used in~\cite{chen2023deep} to transfer the view-complete MvC data into view-complete settings with the missing rate $\eta$ in the range of $[0.1, 0.3, 0.5]$. We repeat all approaches for 5 times and present the mean clustering results in terms of four metrics on four MvC datasets of all methods with missing rates varying from 0.1 to 0.5 in Tab.~\ref{tab:imvc_results}.

From the Tab.~\ref{tab:imvc_results}, we obtain the following observations: 
The proposed method achieves better clustering results than other compared approaches on different datasets with diverse missing rates. For example, our method performs better than the second-best results about 8.71, 3.11, 9.99, and 1.2 percentages measured by ACC, NMI, ARI, and Fscore on the OutdoorScene dataset. Such clustering performance advantages strongly indicate the effectiveness and superiority of the proposed method, which introduces a mask-informed fusion framework with a prior knowledge-assisted contrastive learning manner. The CPSPAN method imputes the view-specific latent features based on the cross-view similarity graph, which is used to find the nearest neighbor to fill missing samples. However, the imputation procedure without the true labels inevitably causes inaccurate missing view imputation, and consequentially degrades the final clustering results. GIGA and the proposed method are both imputation-free kind of approaches, which obtain better clustering results than the CPSPAN method. The MVCAN and SCM are originally designed for view-complete MvC datasets. In our experiments, we directly fill the missing samples with zeros and input MVCAN and SCM to get the clustering results. As seen from the results, MVCAN and SCM can still obtain consideration clustering performance compared with other well-designed incomplete MvC approaches in small missing ratios. The proposed method consistently performs better than MVCAN and SCM on different datasets with diverse missing ratio settings. This indicates that effectively leveraging the observation status of different samples across multiple views is beneficial for aggregating the information from IMvC data.

\subsection{Model Analysis}
In this section, we conduct a series of experiments to analyze the effectiveness of diverse components and different settings in the proposed method. 

\textbf{Effectiveness of mask-informed fusion strategy}:
The mask-informed fusion strategy consists of two parts, i.e., the mask-informed feature fusion and mask-informed graph fusion, which are utilized to aggregate the incomplete multi-view information with the help of data observation status across different views. Here, we formulate two methods called wo-MFF and wo-MGF, which remove the mask-informed feature fusion and mask-informed graph fusion from the model, respectively. As shown in Tab.~\ref{tab:imvc_ab}, the clustering performance of the proposed method consistently outperforms than wo-MFF and wo-MGF methods, indicating the effectiveness of the proposed mask-informed fusion strategy.

\textbf{Effectiveness of prior knowledge-assisted contrastive learning}:
The prior knowledge-assisted contrastive learning is a key component of the proposed method, which leverages the neighbor information captured from diverse views to boost the learning process of the view-common feature representation for clustering with a weighted contrastive loss. To study its effectiveness, we first remove the prior knowledge-assisted contrastive learning from our model and call it wo-WCL. As shown in Tab.~\ref{tab:imvc_ab}, the clustering performance of the wo-WCL method significantly drops compared to the proposed model. In addition, we evaluate the clustering performance of the proposed model with only the prior knowledge-assisted contrastive loss without the feature reconstruction part and term it the wo-Rec method. From the results in Tab.~\ref{tab:imvc_ab}, we observe that only with the prior knowledge-assisted contrastive loss can not achieve that optimal clustering performance. Thus, the feature reconstruction and prior knowledge-assisted contrastive learning are both important for the feature representation learning in the proposed method. Finally, we compare our weighted contrastive loss with the widely used contrastive loss~\cite{chen2020simple} and decoupled contrastive loss~\cite{yeh2022decoupled}, which are marked as w-CL and w-DCL methods in Tab.~\ref{tab:imvc_ab}. From the results, we find that 1) our weighted contrastive loss performs better than contrastive loss and decoupled contrastive loss in most cases, which indicates the weighted contrastive loss is suitable to utilize the prior knowledge of the incomplete multi-view data. 2) The weighted contrastive loss and decoupled contrastive loss achieve better results than contrastive loss indicating that the weighted contrastive loss and decoupled contrastive loss are less sensitive to the negative-positive-coupling issues~\cite{yeh2022decoupled}.

\begin{figure*}[!htbp]
	\centering
	\subfigure[MSRCV]{
		\includegraphics[width=0.23\textwidth]{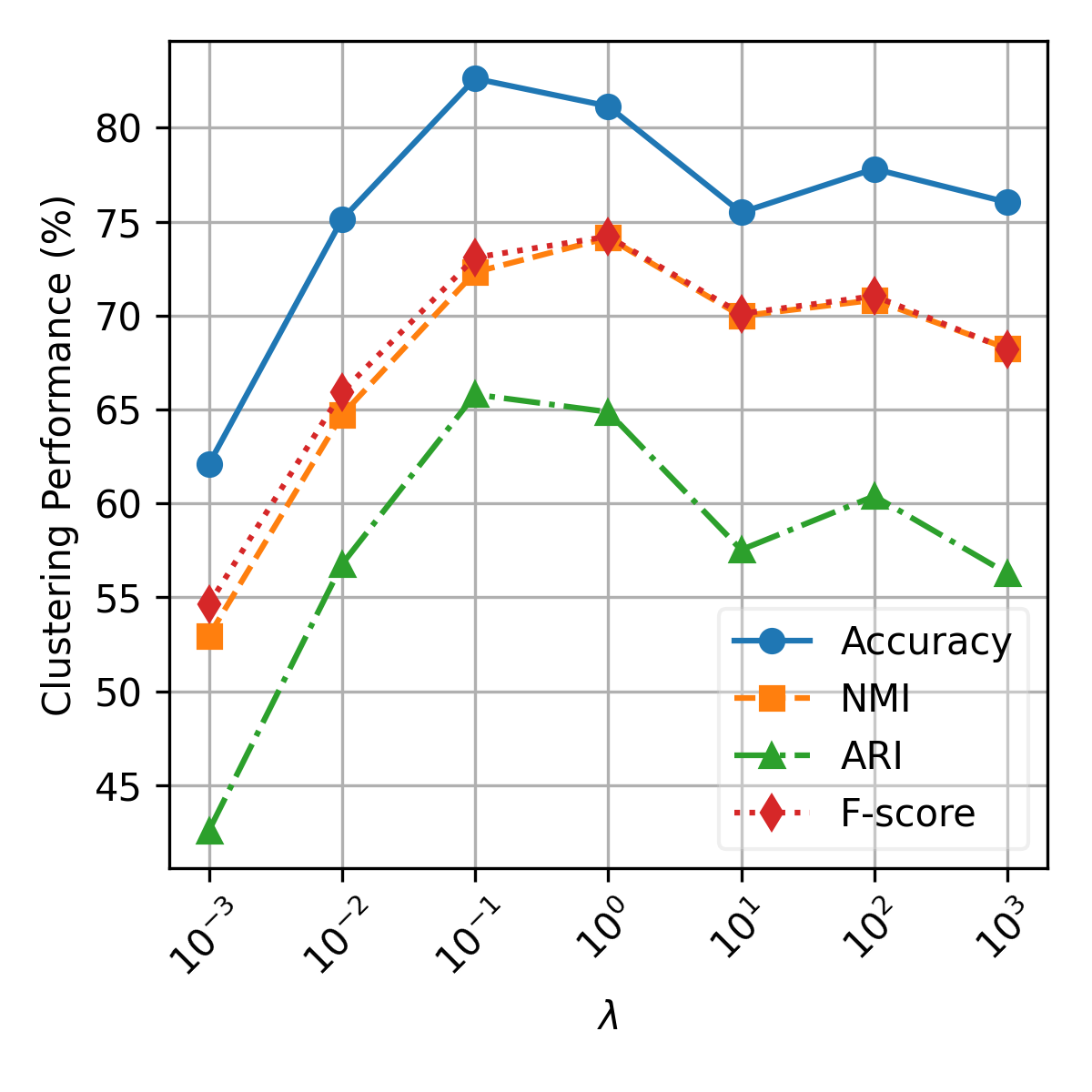}\label{fig:MSRCV_ps}}
	\subfigure[CUB]{
		\includegraphics[width=0.23\textwidth]{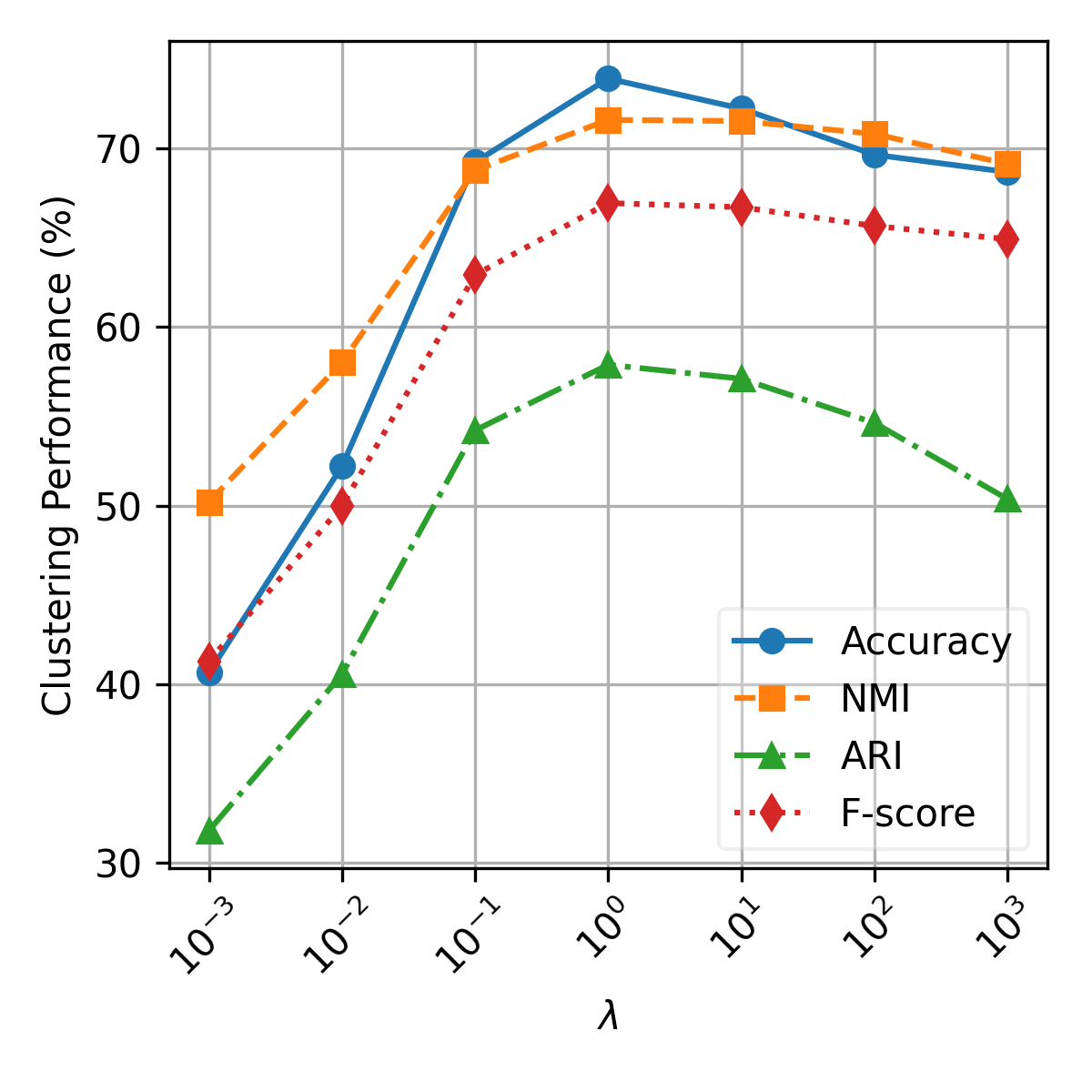}\label{fig:CUB_ps}}
	\subfigure[nuswide]{
		\includegraphics[width=0.23\textwidth]{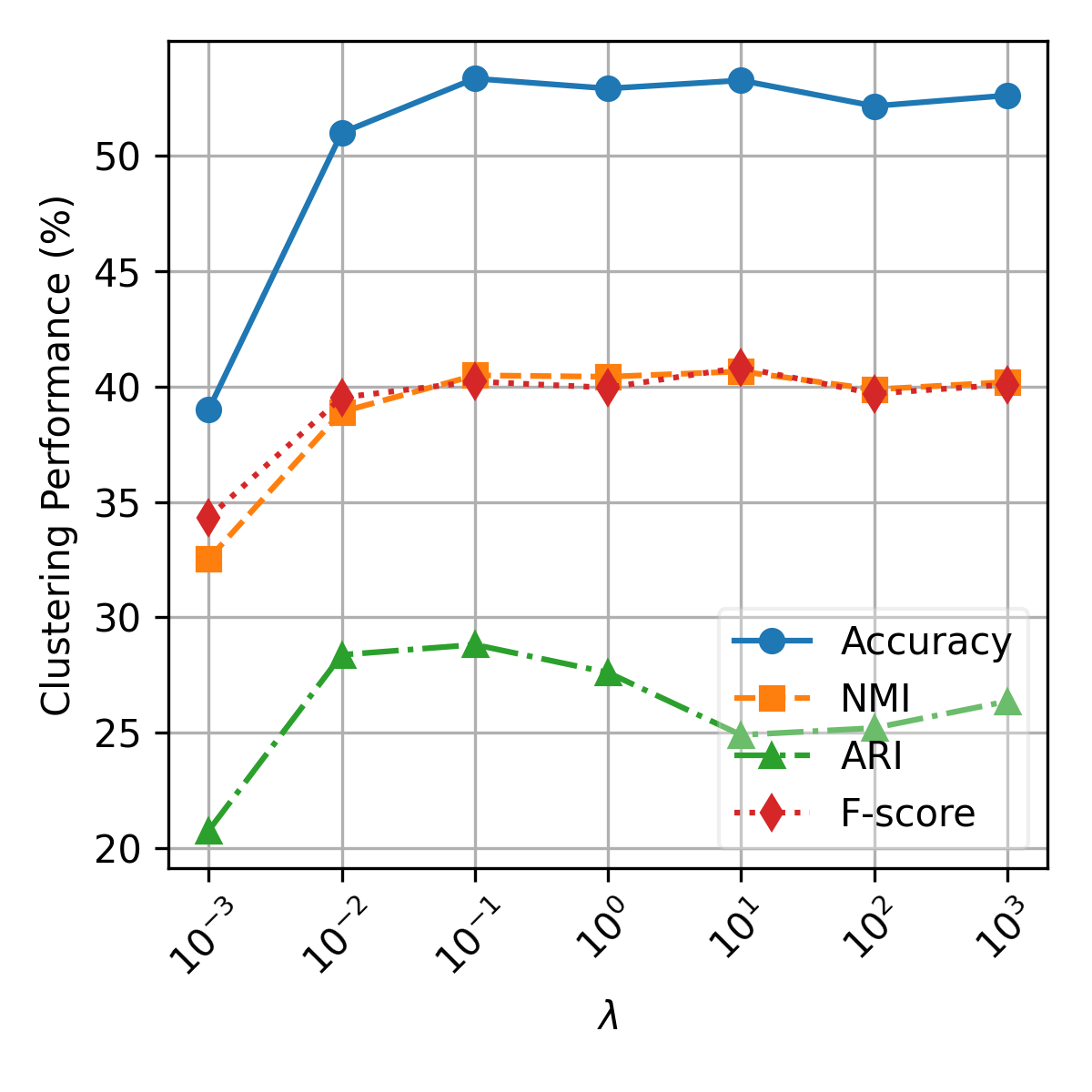}\label{fig:nuswide_ps}}
	\subfigure[OutdoorScene]{
		\includegraphics[width=0.23\textwidth]{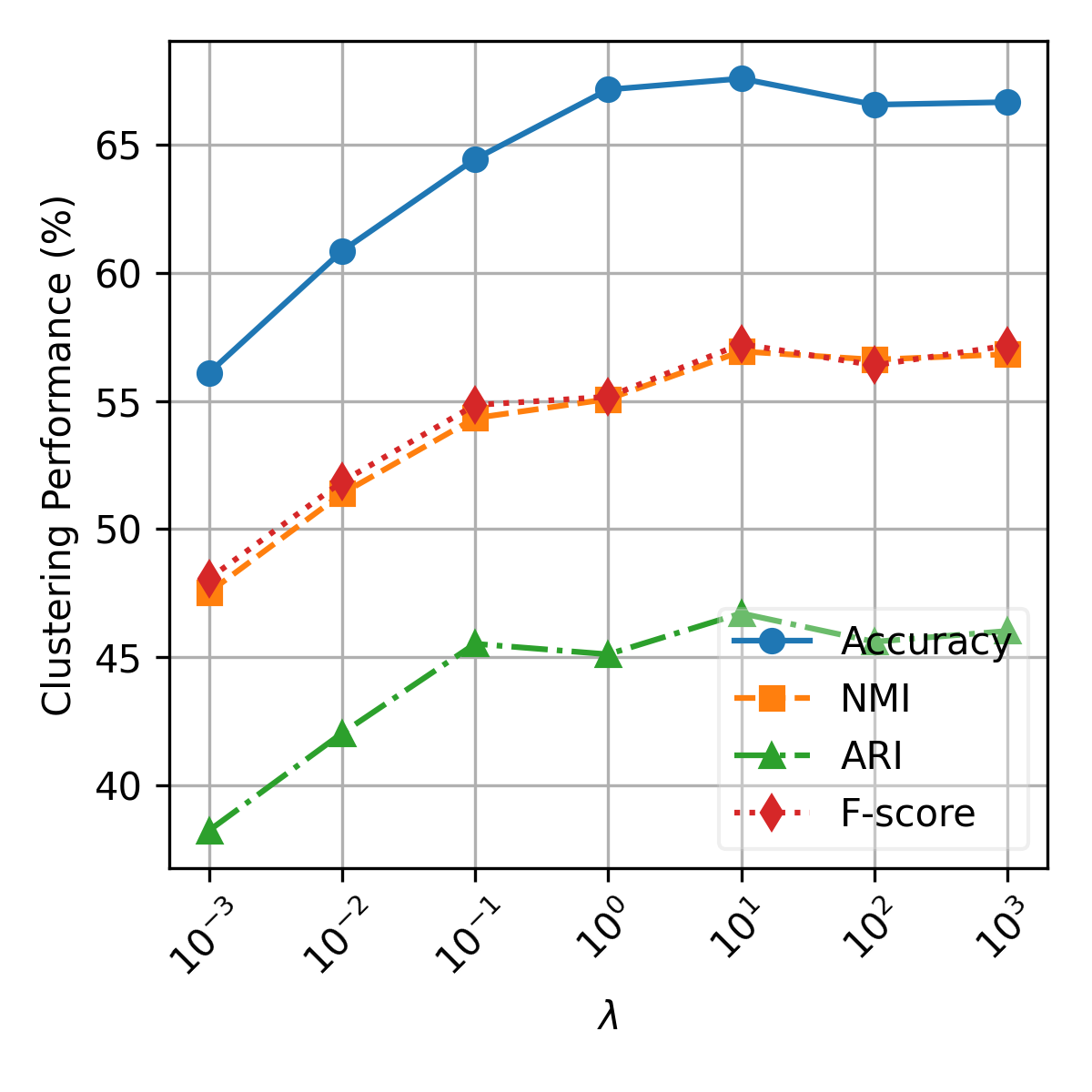}\label{fig:OutdoorScene_ps}}
	\caption{The parameter sensitivity of the proposed method on four multi-view datasets in terms of ACC, NMI, ARI, and Fscore, respectively.}
	\label{fig:ps_ap}
\end{figure*}

\begin{figure*}[!htbp]
	\centering
	\subfigure[MSRCV]{
		\includegraphics[width=0.23\textwidth]{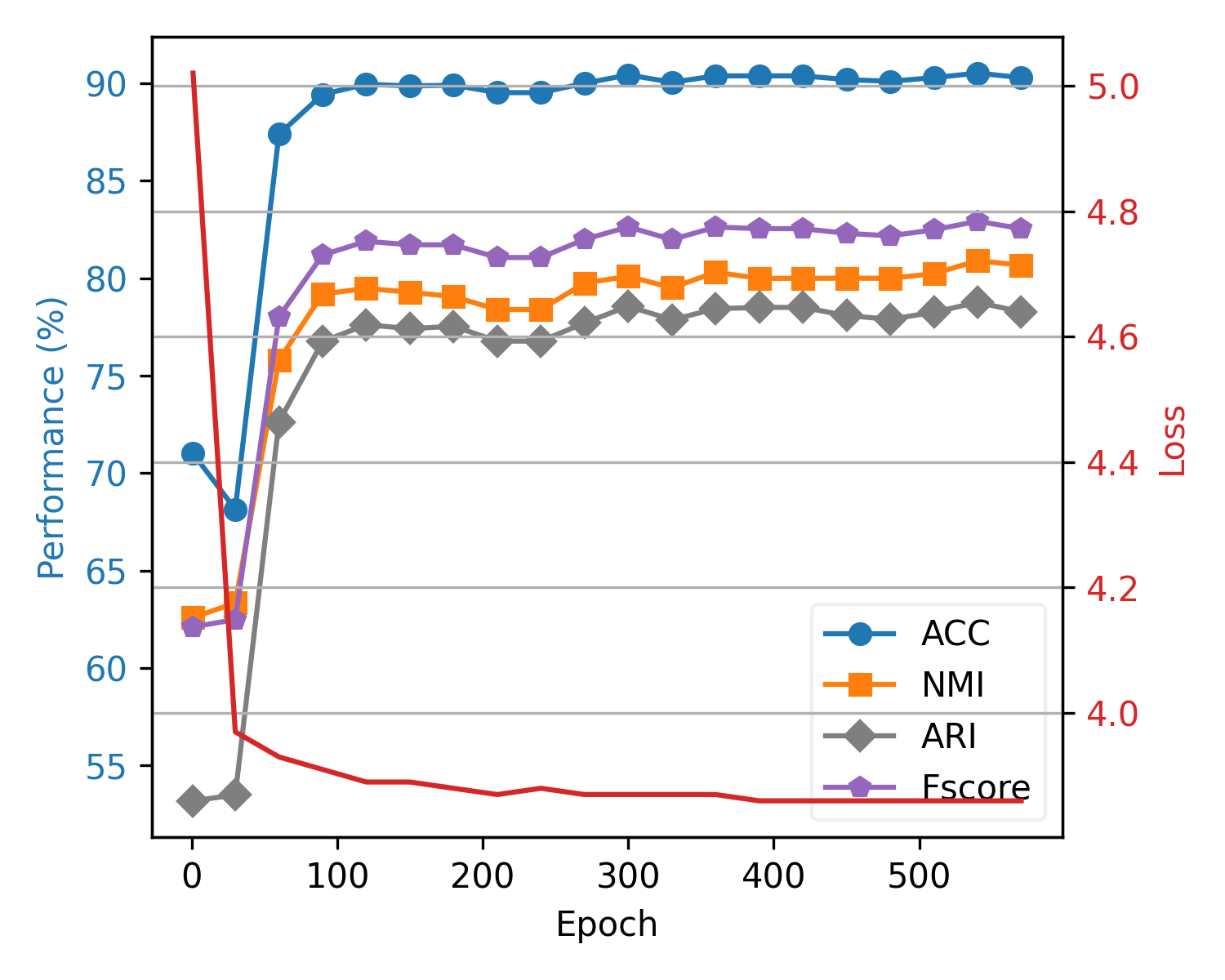}\label{fig:MSRCV_loss}}
	\subfigure[CUB]{
		\includegraphics[width=0.23\textwidth]{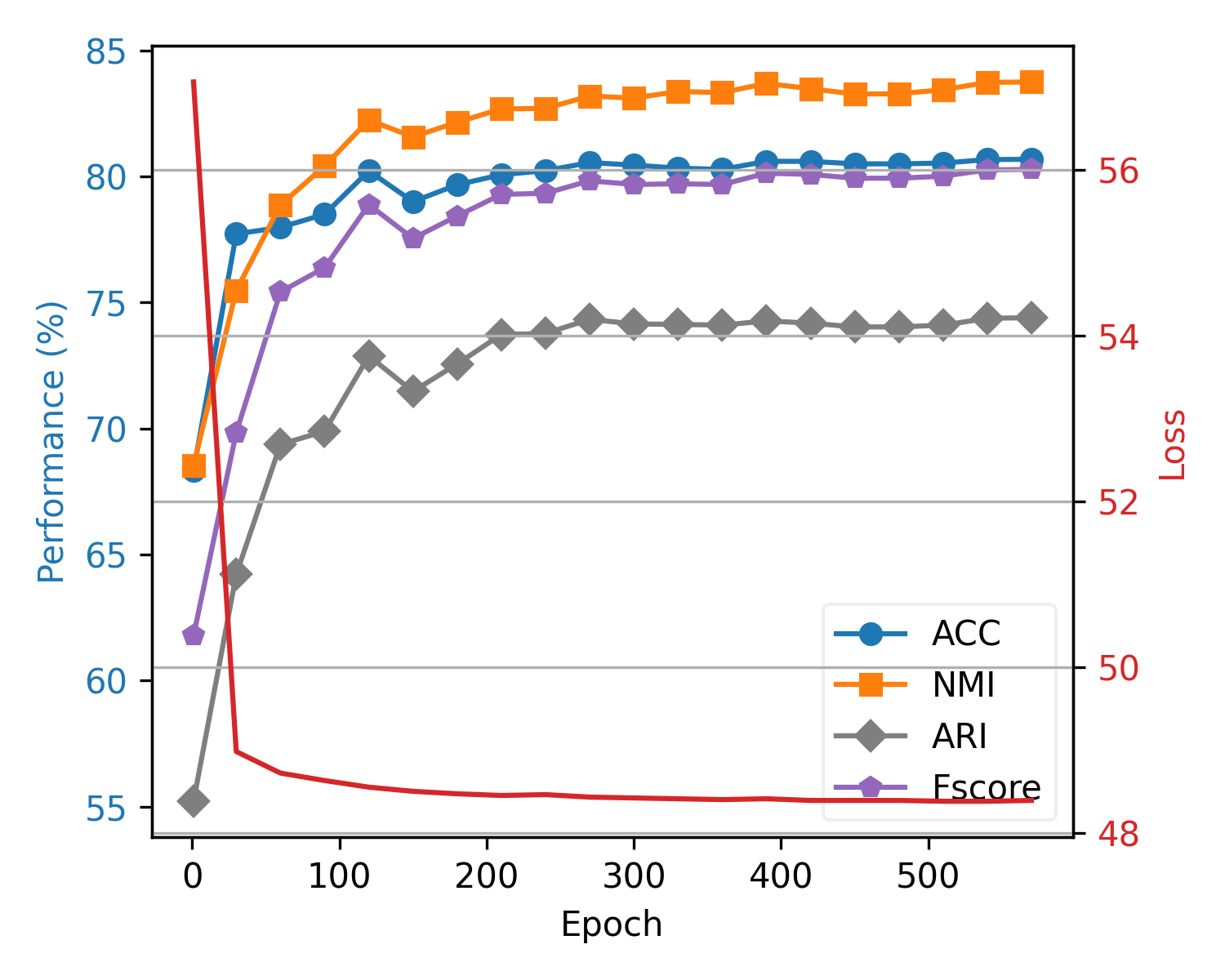}\label{fig:CUB_loss}}
	\subfigure[nuswide]{
		\includegraphics[width=0.23\textwidth]{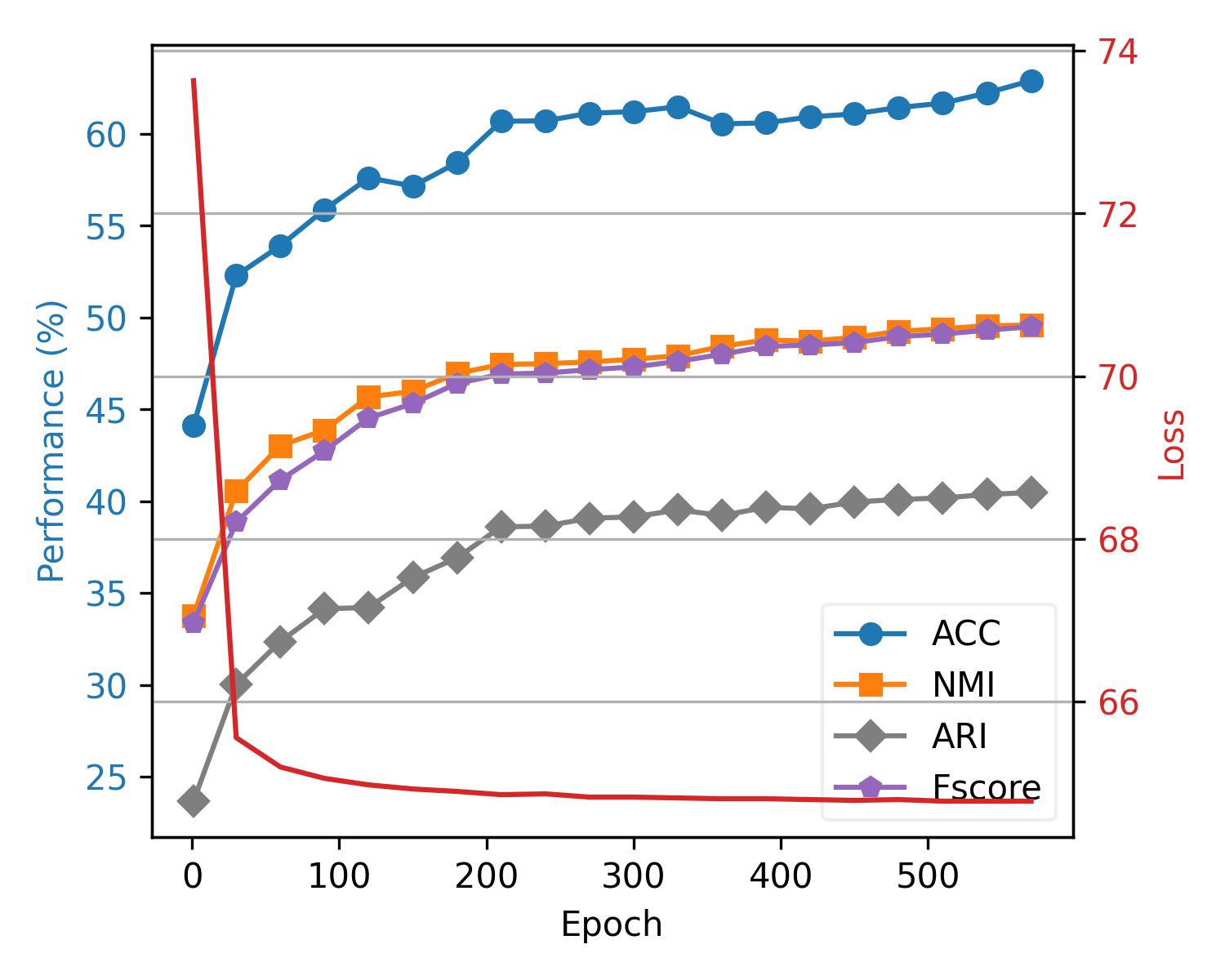}\label{fig:nuswide_loss}}
	\subfigure[OutdoorScene]{
		\includegraphics[width=0.23\textwidth]{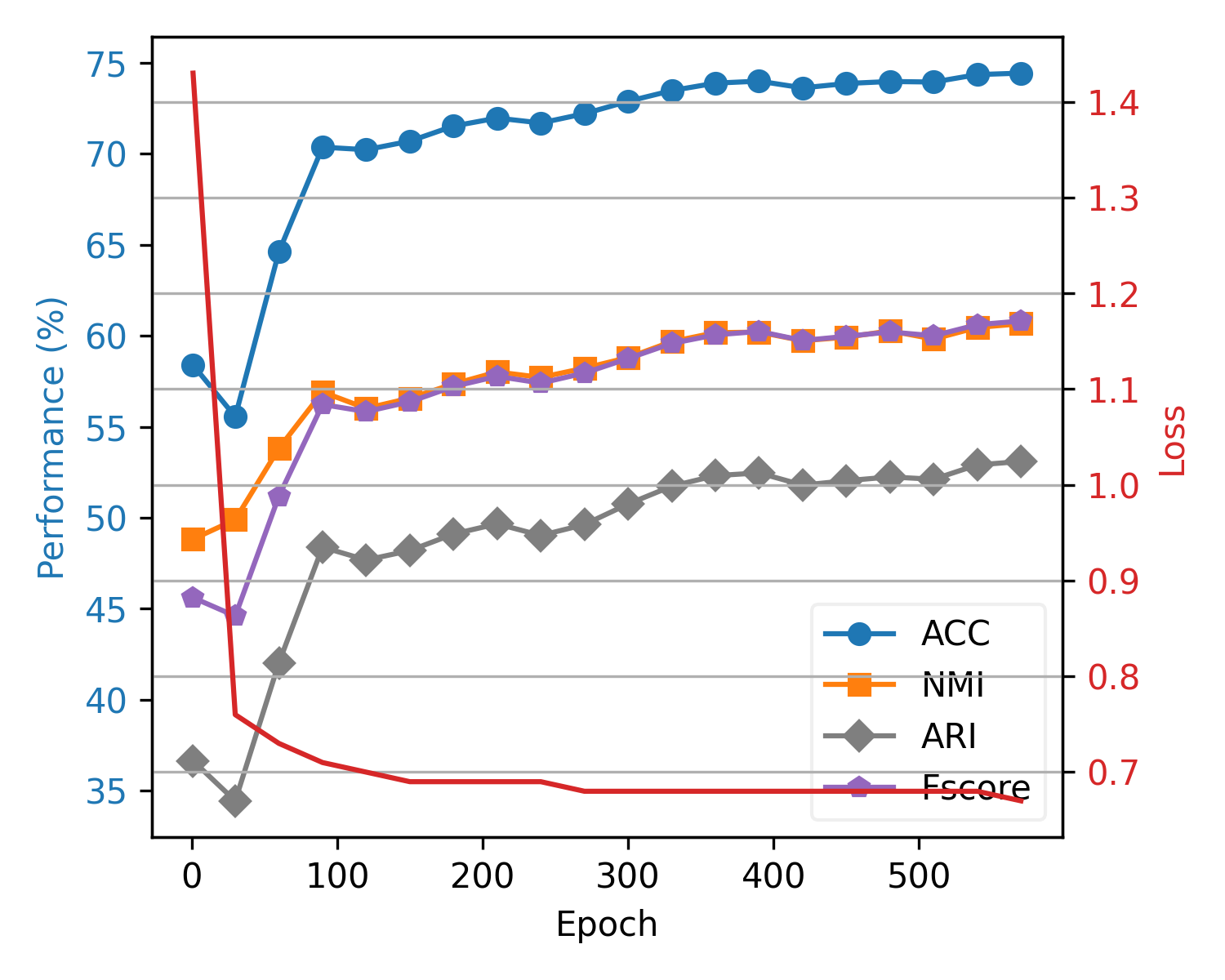}\label{fig:OutdoorScene_loss}}
	\caption{The convergence analysis of the proposed method on four benchmark multi-view datasets in terms of ACC, NMI, ARI, and Fscore, respectively.}
	\label{fig:loss}
\end{figure*}

\begin{figure*}[!htbp]
	\centering
	\subfigure[$t$=1]{
		\includegraphics[width=0.23\textwidth]{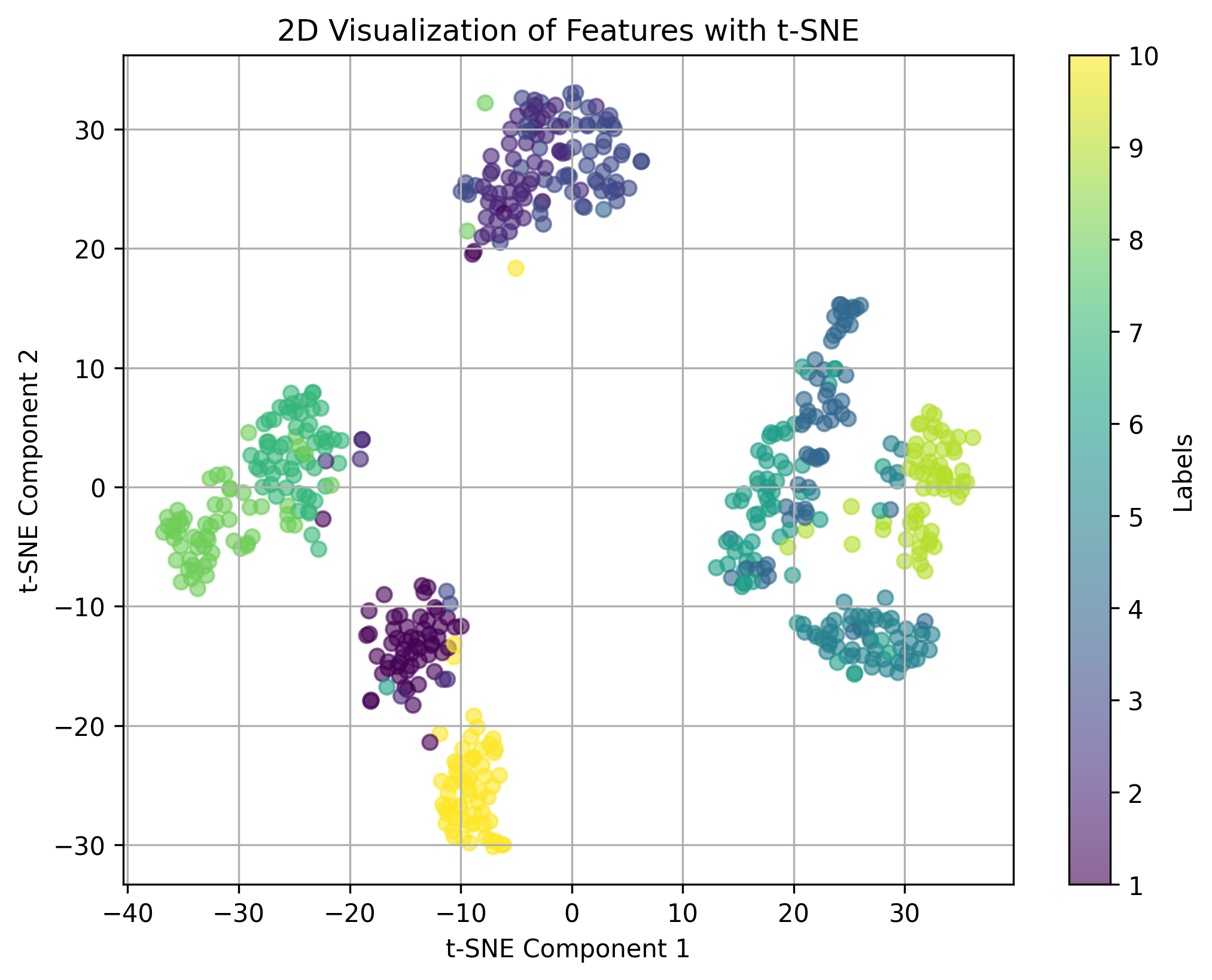}\label{fig:CUB_1}}
	\subfigure[$t$=50]{
		\includegraphics[width=0.23\textwidth]{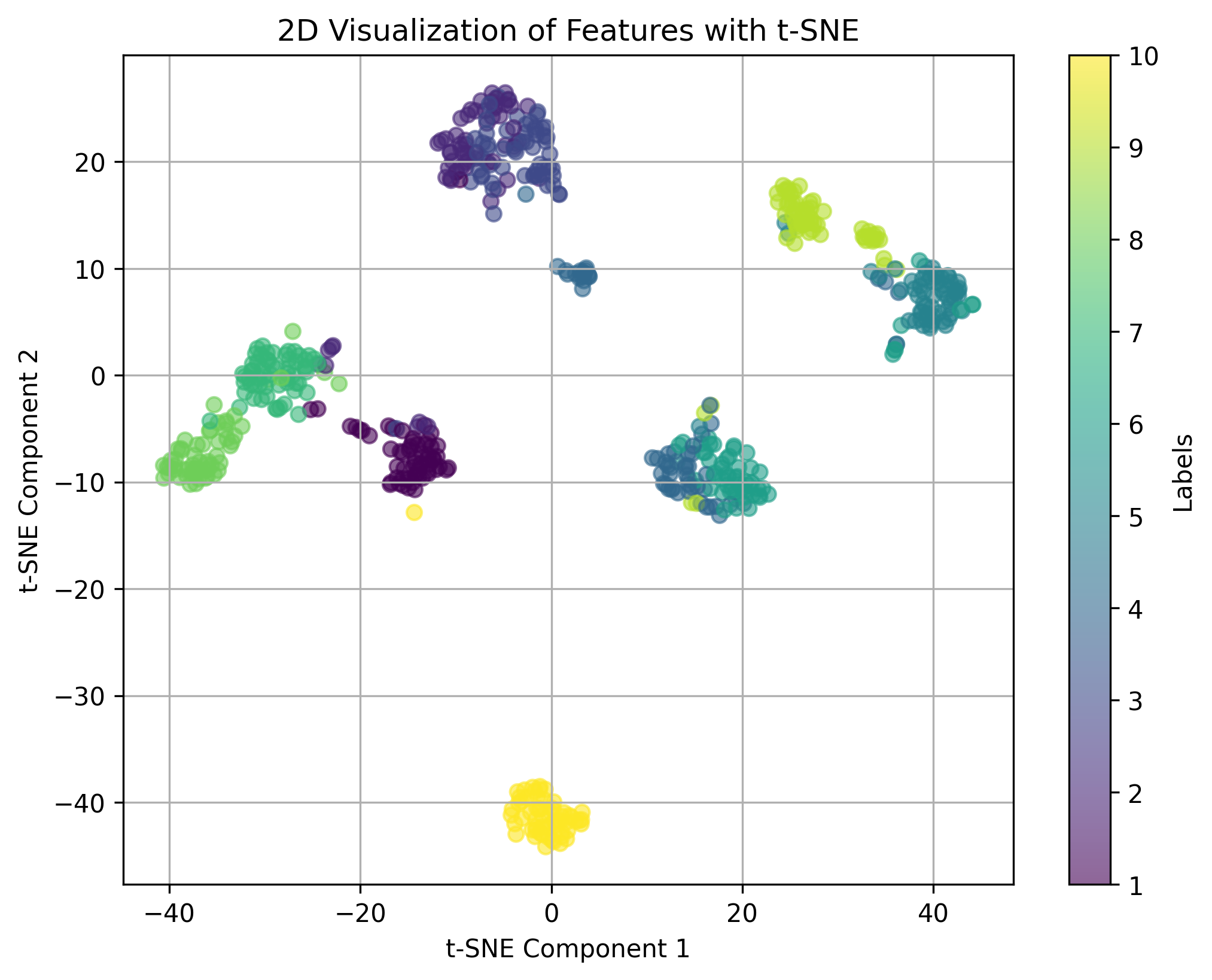}\label{fig:CUB_50}}
	\subfigure[$t$=100]{
		\includegraphics[width=0.23\textwidth]{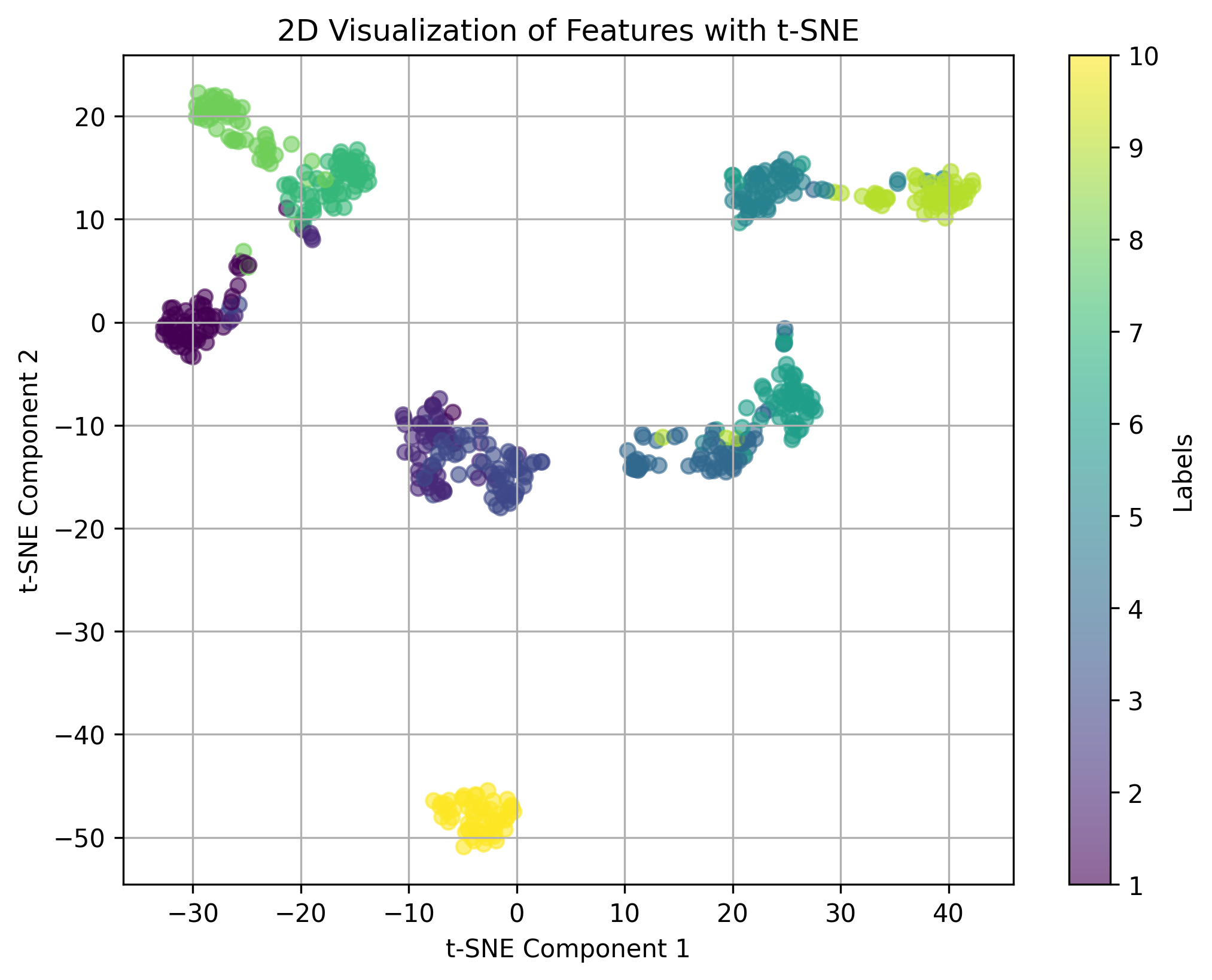}\label{fig:CUB_100}}
	\subfigure[$t$=200]{
		\includegraphics[width=0.23\textwidth]{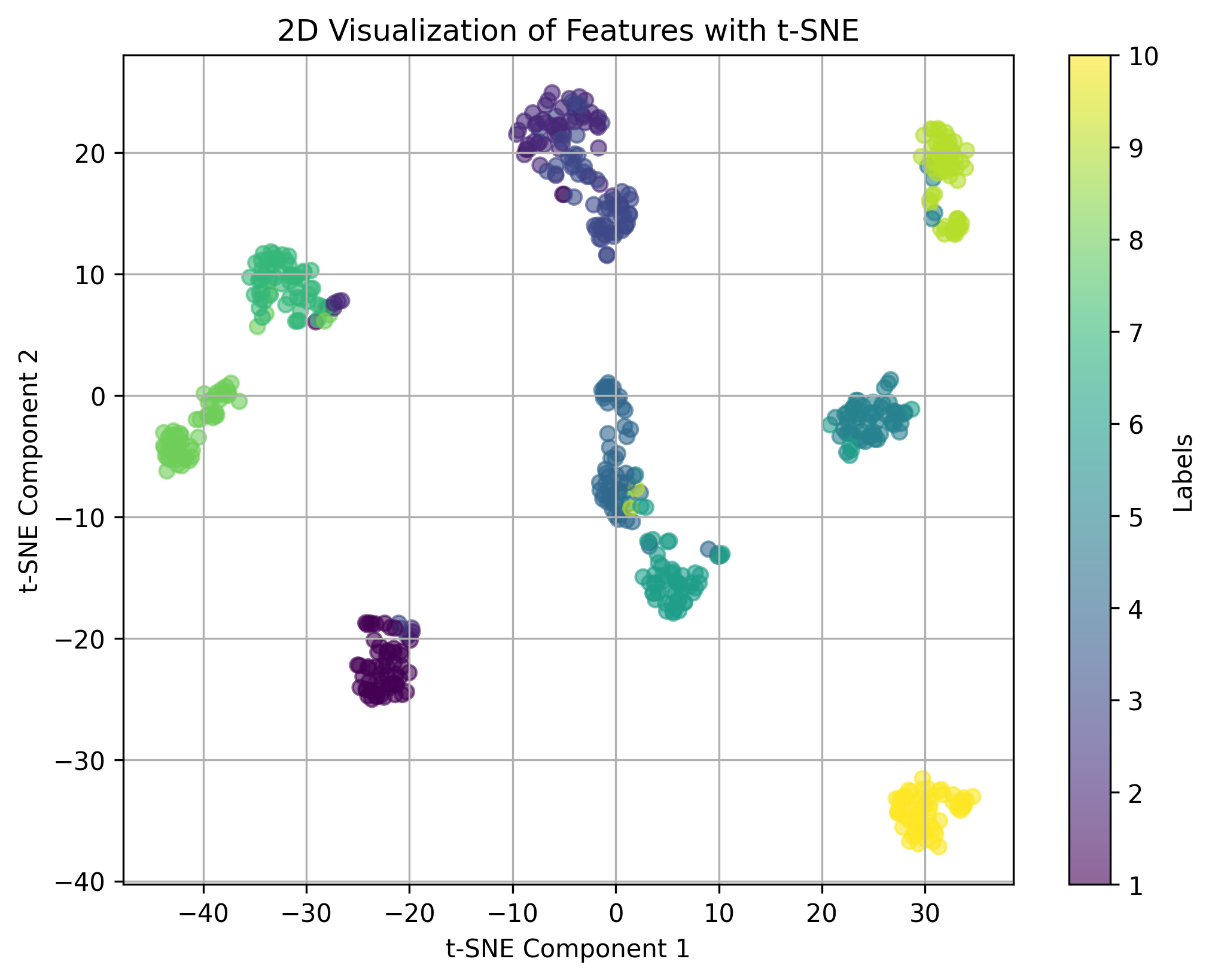}\label{fig:CUB_200}}
	\caption{The visualization results of the proposed method under 1, 50, 100 and 200 epochs on CUB dataset.}
	\label{fig:CUB_tsne}
\end{figure*}

\begin{figure*}[!htbp]
	\centering
	\subfigure[$t$=1]{
		\includegraphics[width=0.23\textwidth]{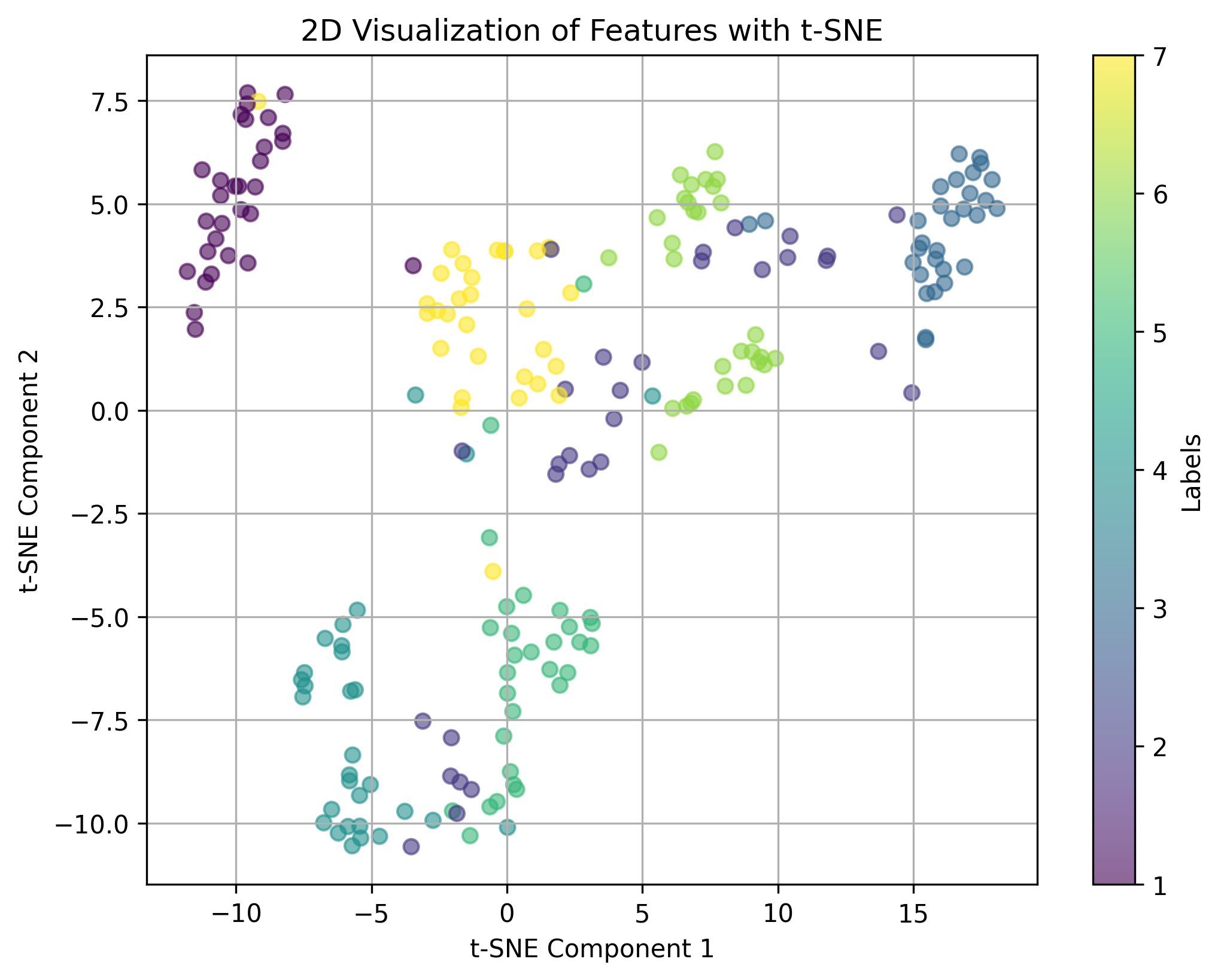}\label{fig:MSRCV_1}}
	\subfigure[$t$=50]{
		\includegraphics[width=0.23\textwidth]{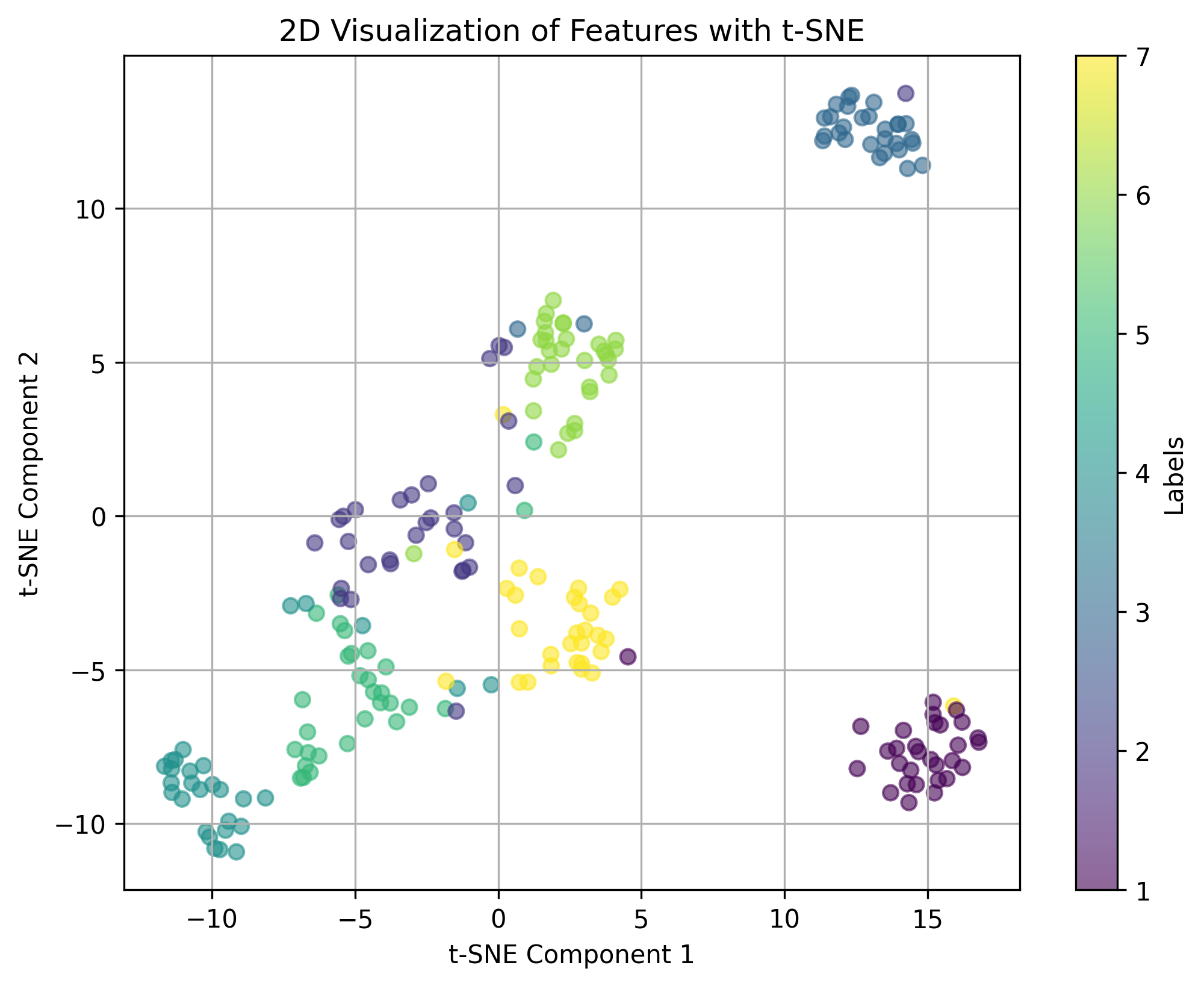}\label{fig:MSRCV_50}}
	\subfigure[$t$=100]{
		\includegraphics[width=0.23\textwidth]{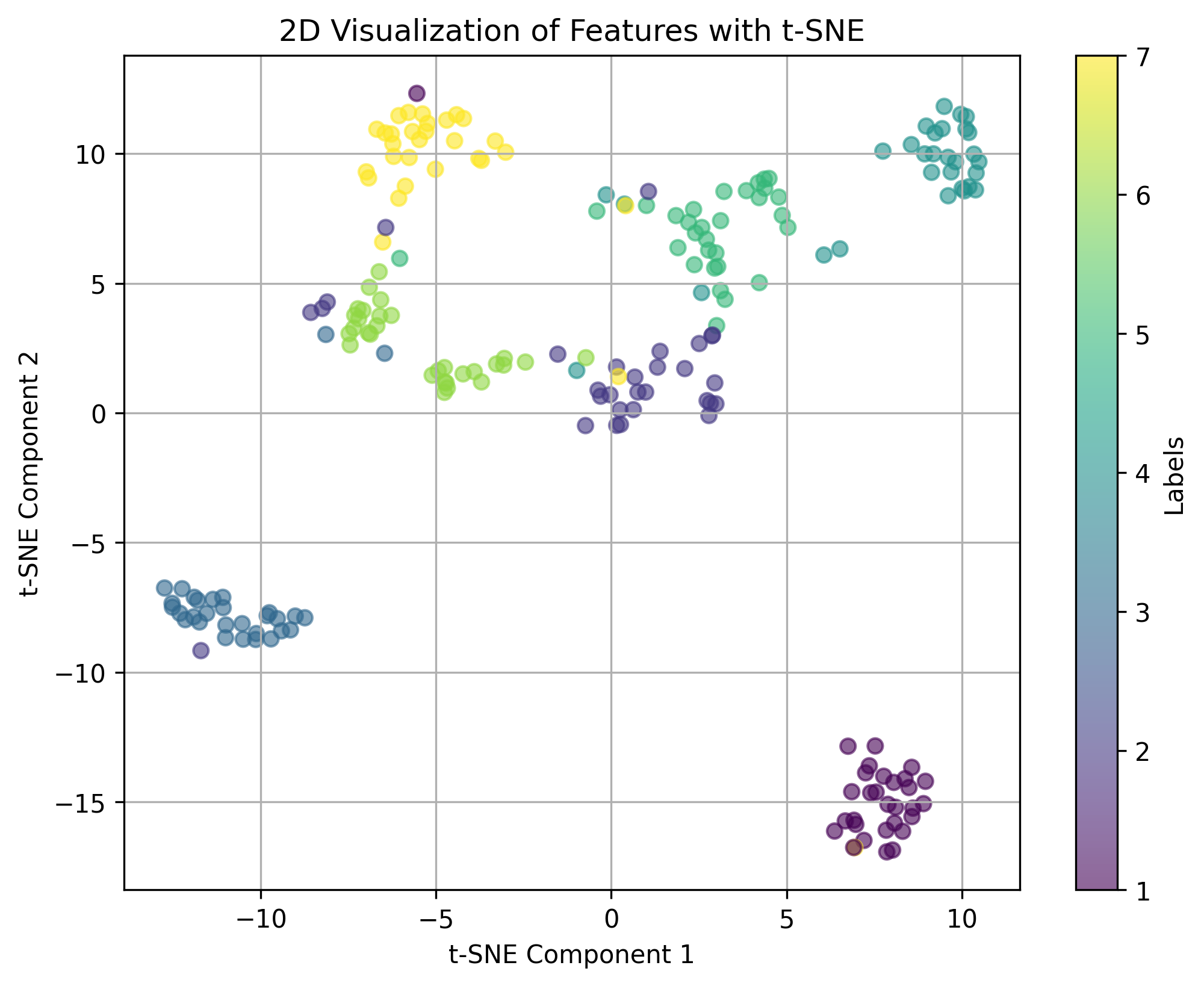}\label{fig:MSRCV_100}}
	\subfigure[$t$=200]{
		\includegraphics[width=0.23\textwidth]{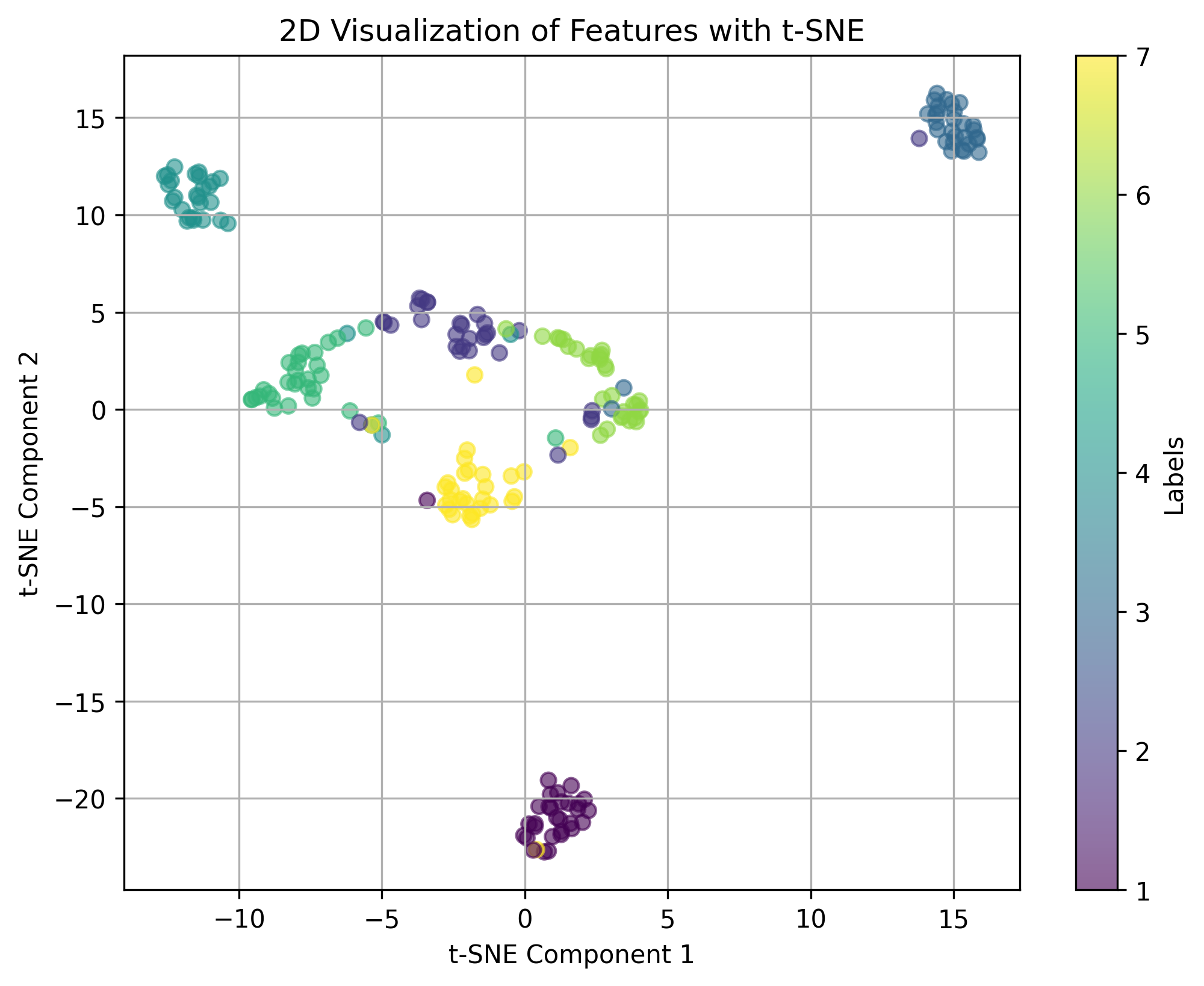}\label{fig:MSRCV_200}}
	\caption{The visualization results of the proposed method under 1, 50, 100 and 200 epochs on MSRCV dataset.}
	\label{fig:MSRCV_tsne}
\end{figure*}

\begin{figure*}[!htbp]
	\centering
	\subfigure[$t$=1]{
		\includegraphics[width=0.23\textwidth]{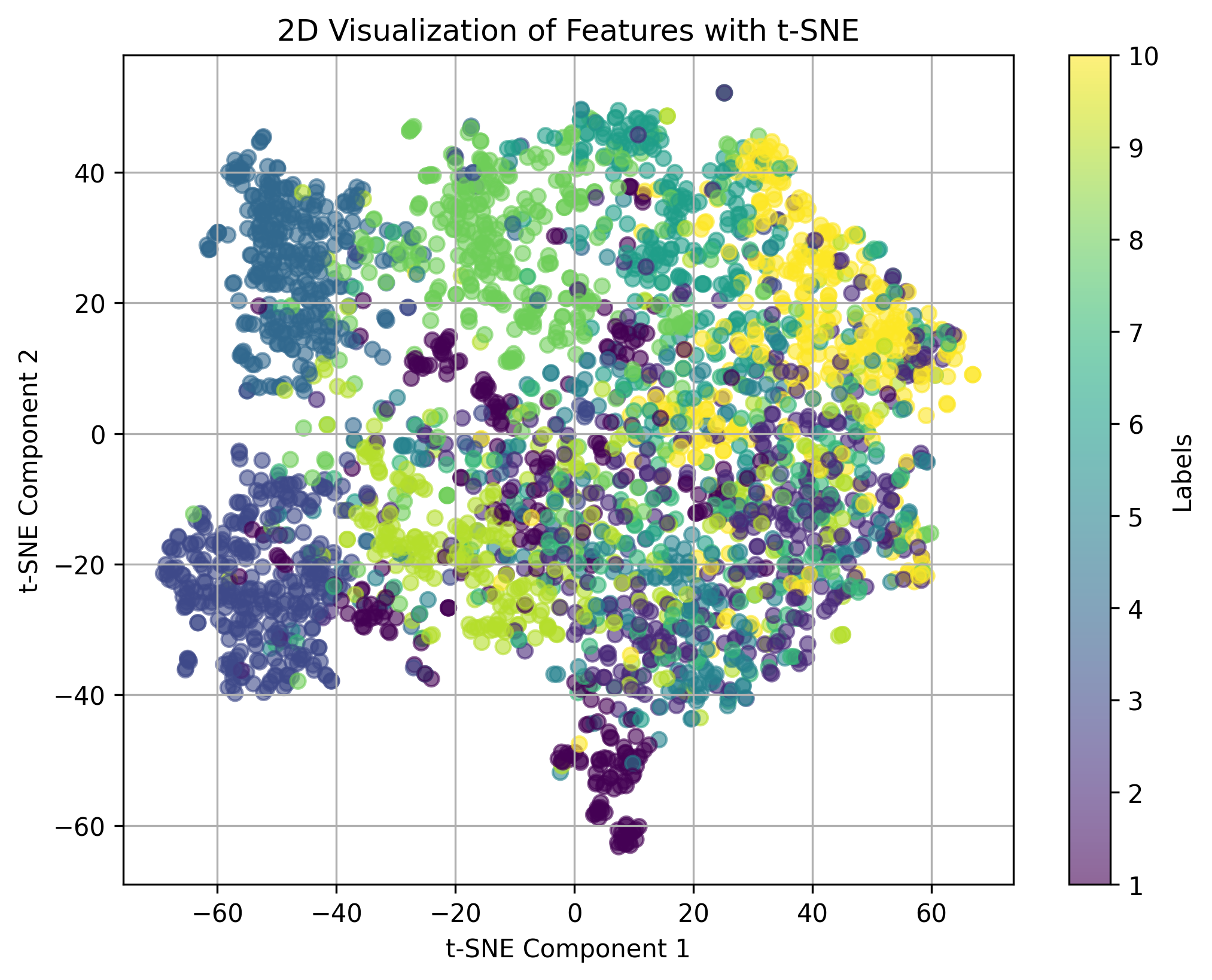}\label{fig:nuswide_1}}
	\subfigure[$t$=50]{
		\includegraphics[width=0.23\textwidth]{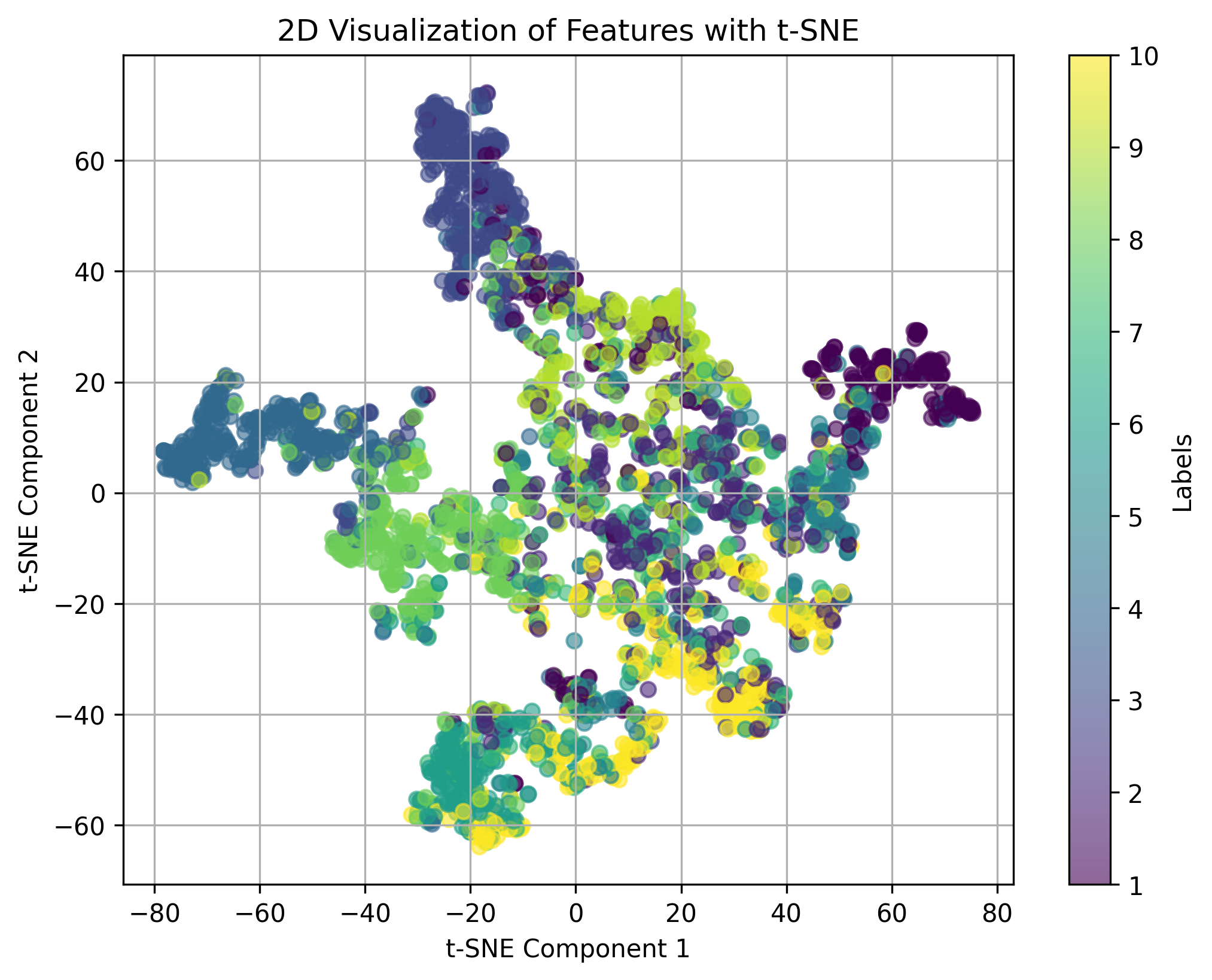}\label{fig:nuswide_50}}
	\subfigure[$t$=100]{
		\includegraphics[width=0.23\textwidth]{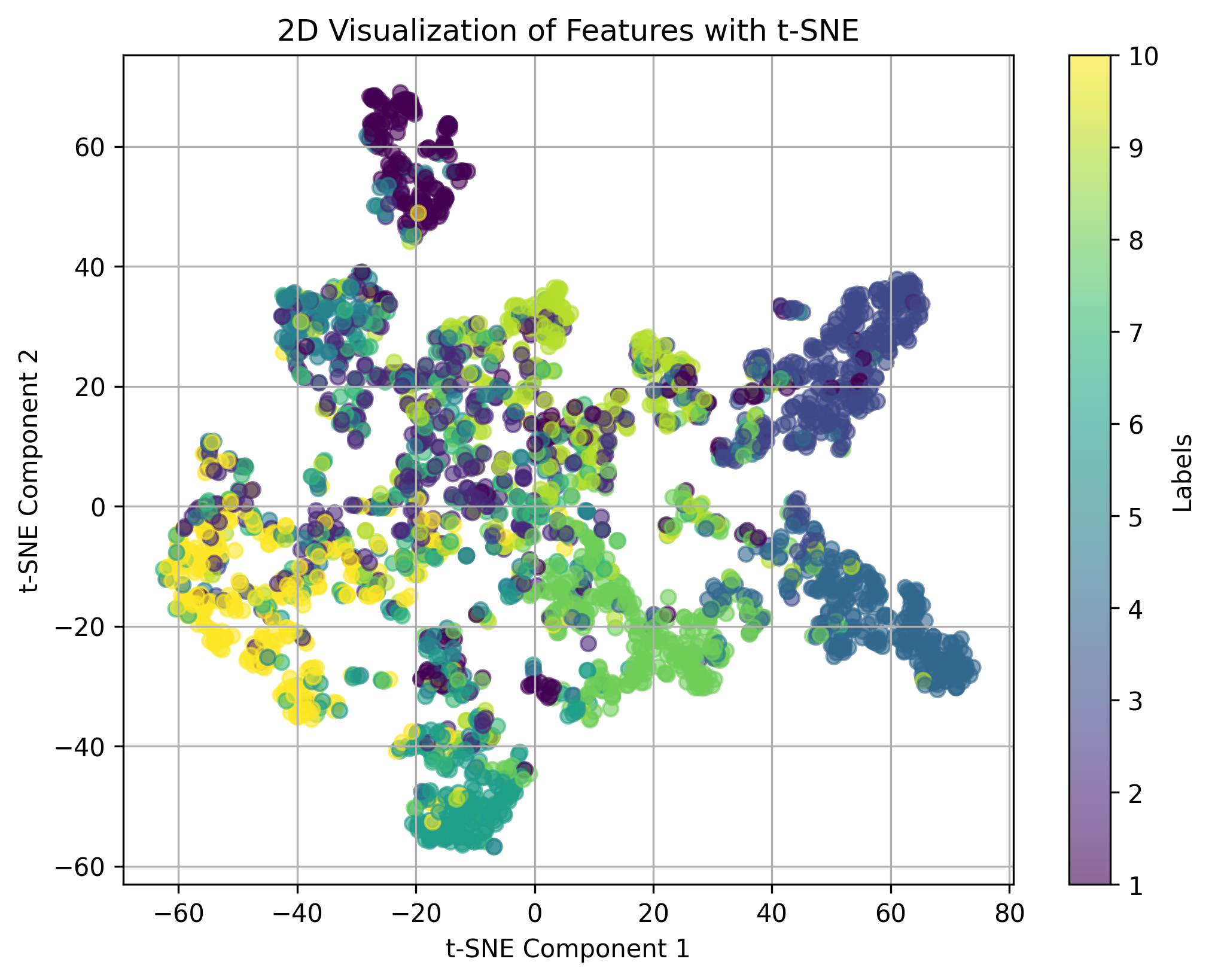}\label{fig:nuswide_100}}
	\subfigure[$t$=200]{
		\includegraphics[width=0.23\textwidth]{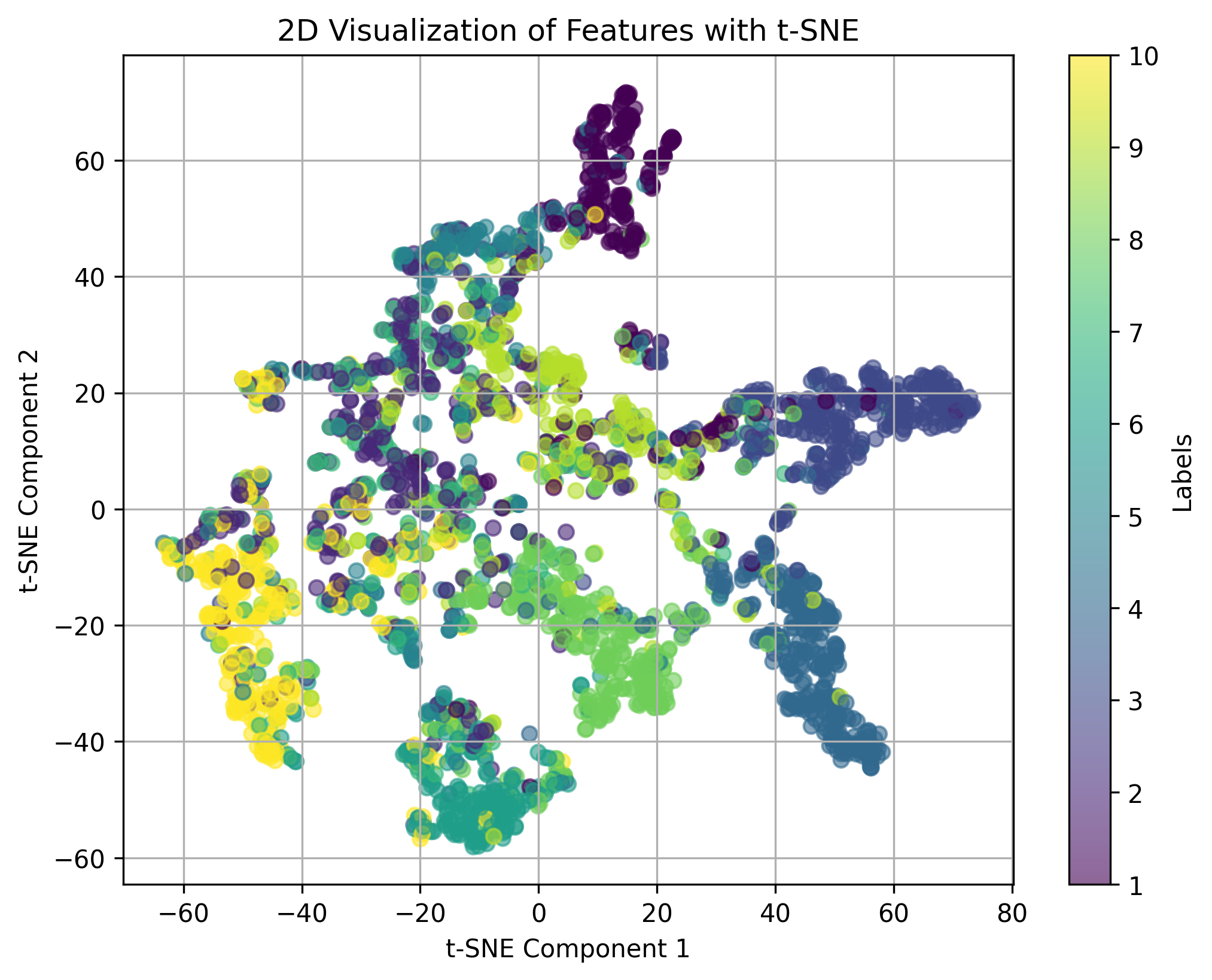}\label{fig:nuswide_200}}
	\caption{The visualization results of the proposed method under 1, 50, 100 and 200 epochs on nuswide dataset.}
	\label{fig:nuswide_tsne}
\end{figure*}

\begin{figure*}[!htbp]
	\centering
	\subfigure[$t$=1]{
		\includegraphics[width=0.23\textwidth]{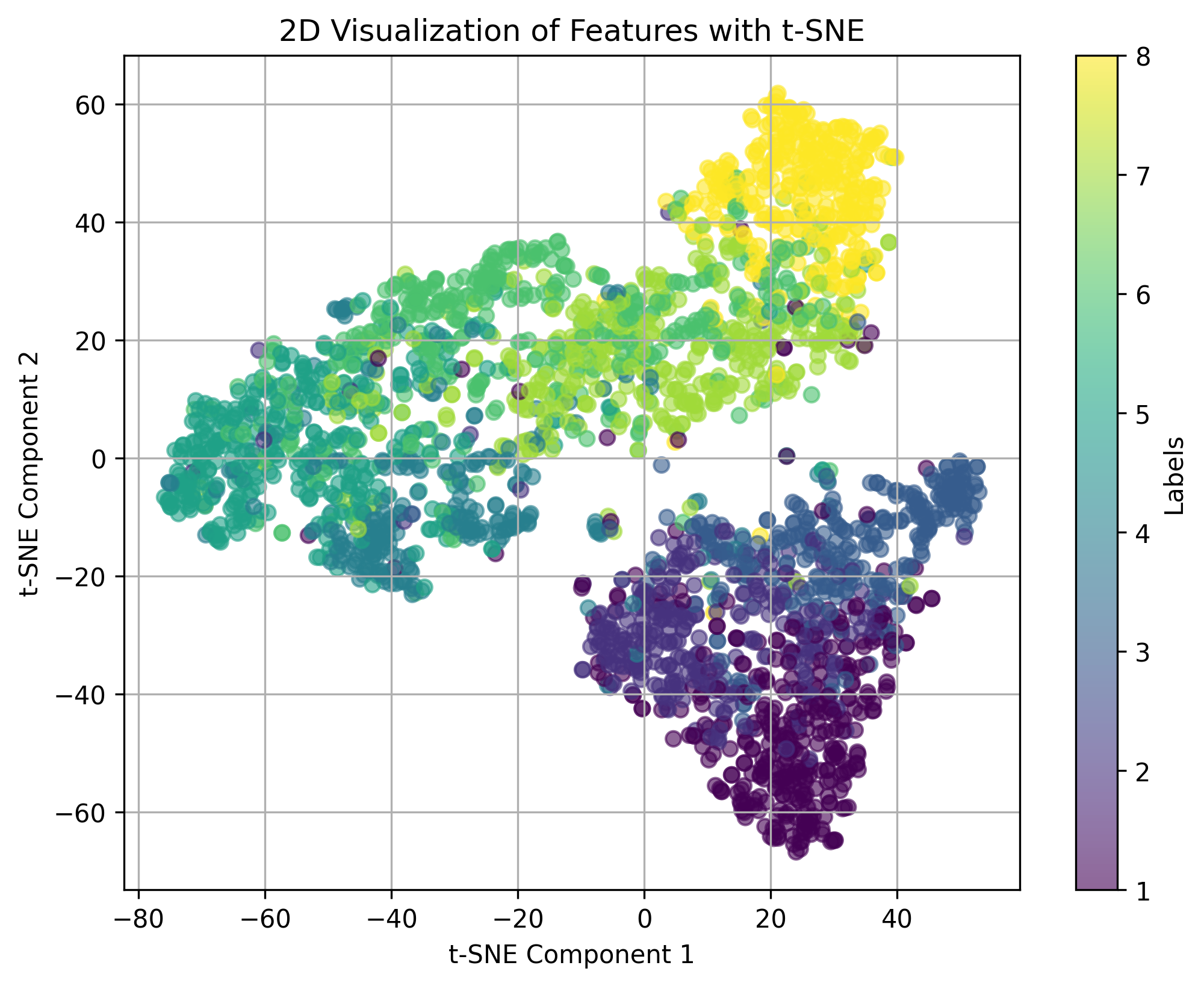}\label{fig:OutdoorScene_1}}
	\subfigure[$t$=50]{
		\includegraphics[width=0.23\textwidth]{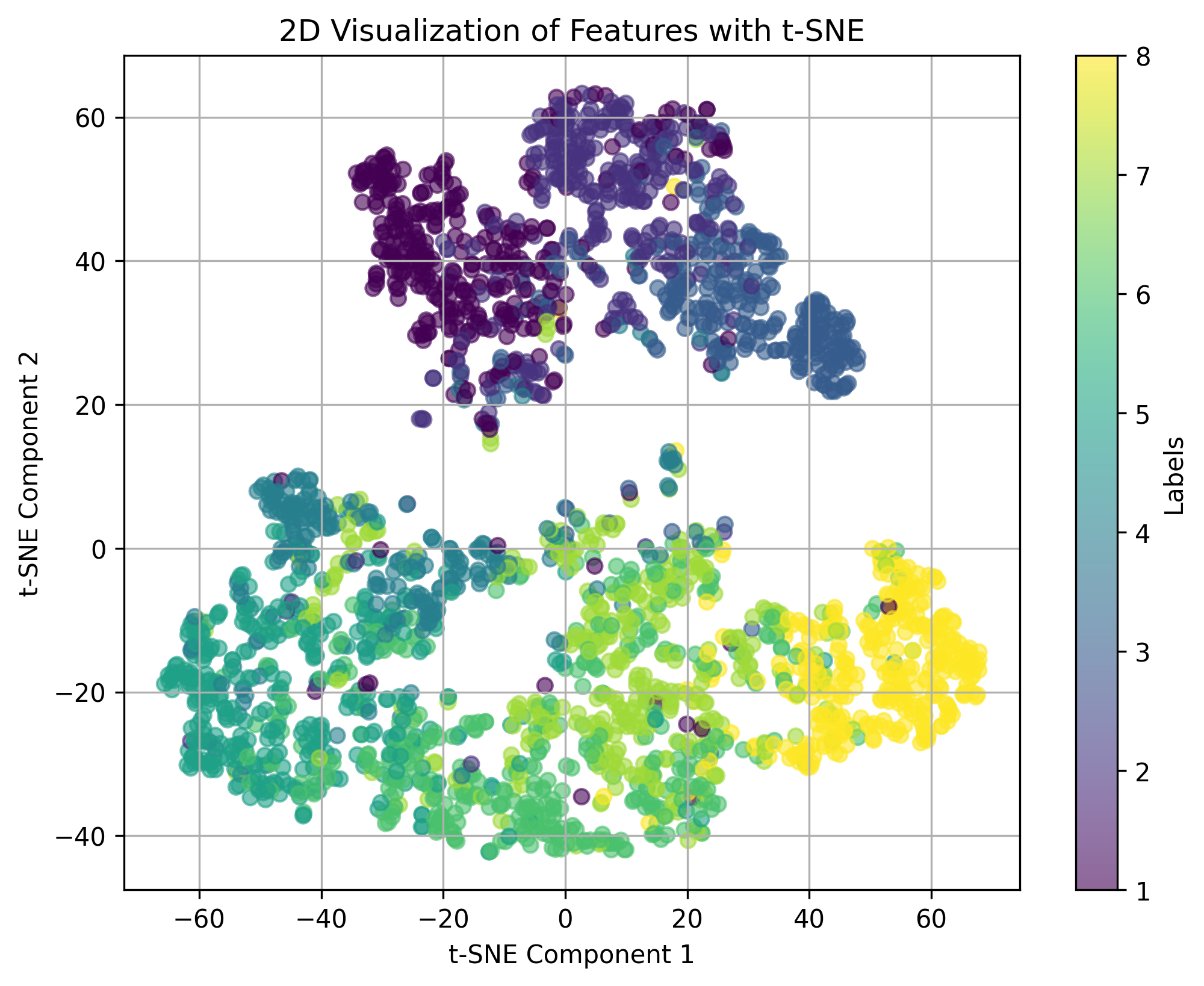}\label{fig:OutdoorScene_50}}
	\subfigure[$t$=100]{
		\includegraphics[width=0.23\textwidth]{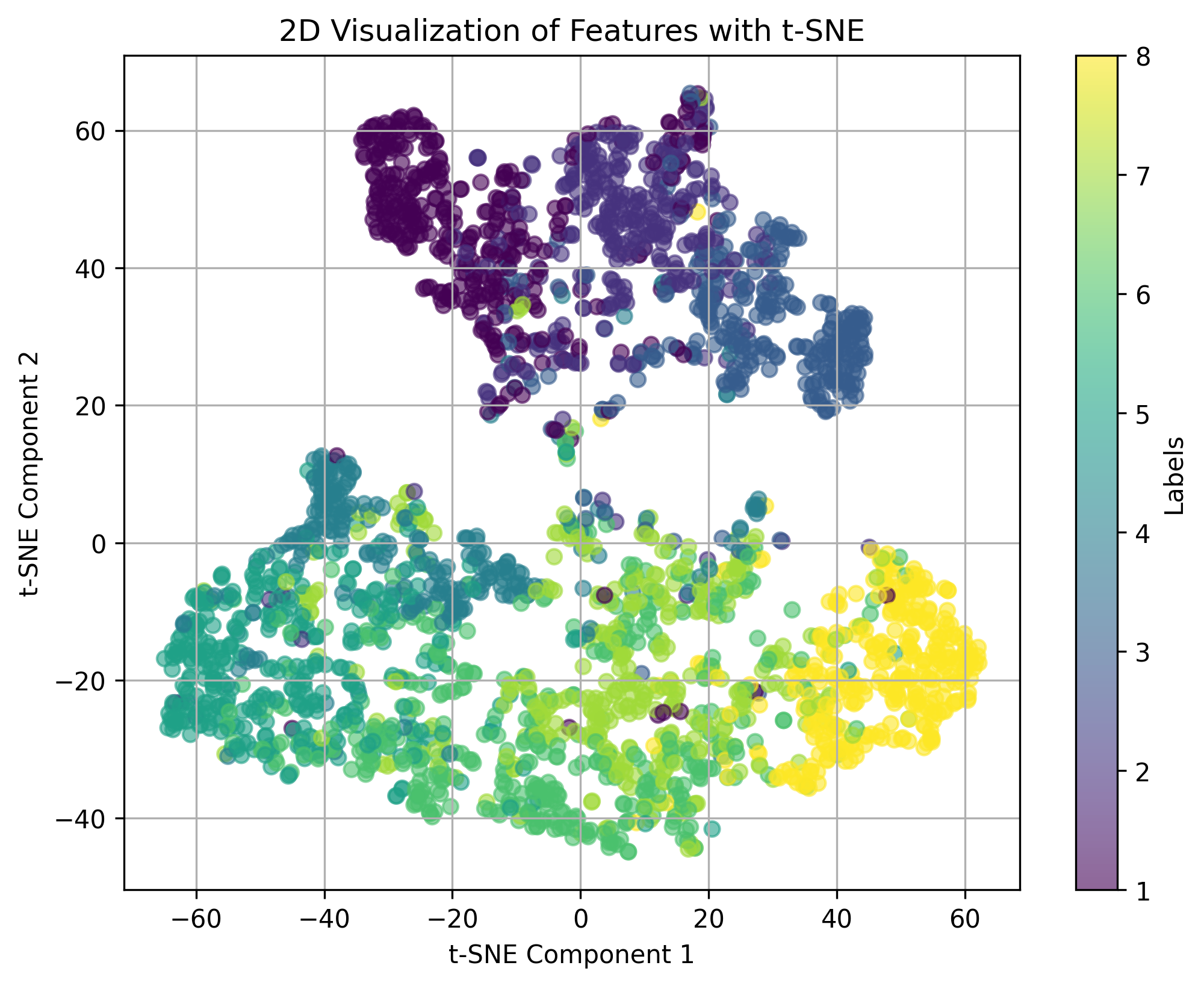}\label{fig:OutdoorScene_100}}
	\subfigure[$t$=200]{
		\includegraphics[width=0.23\textwidth]{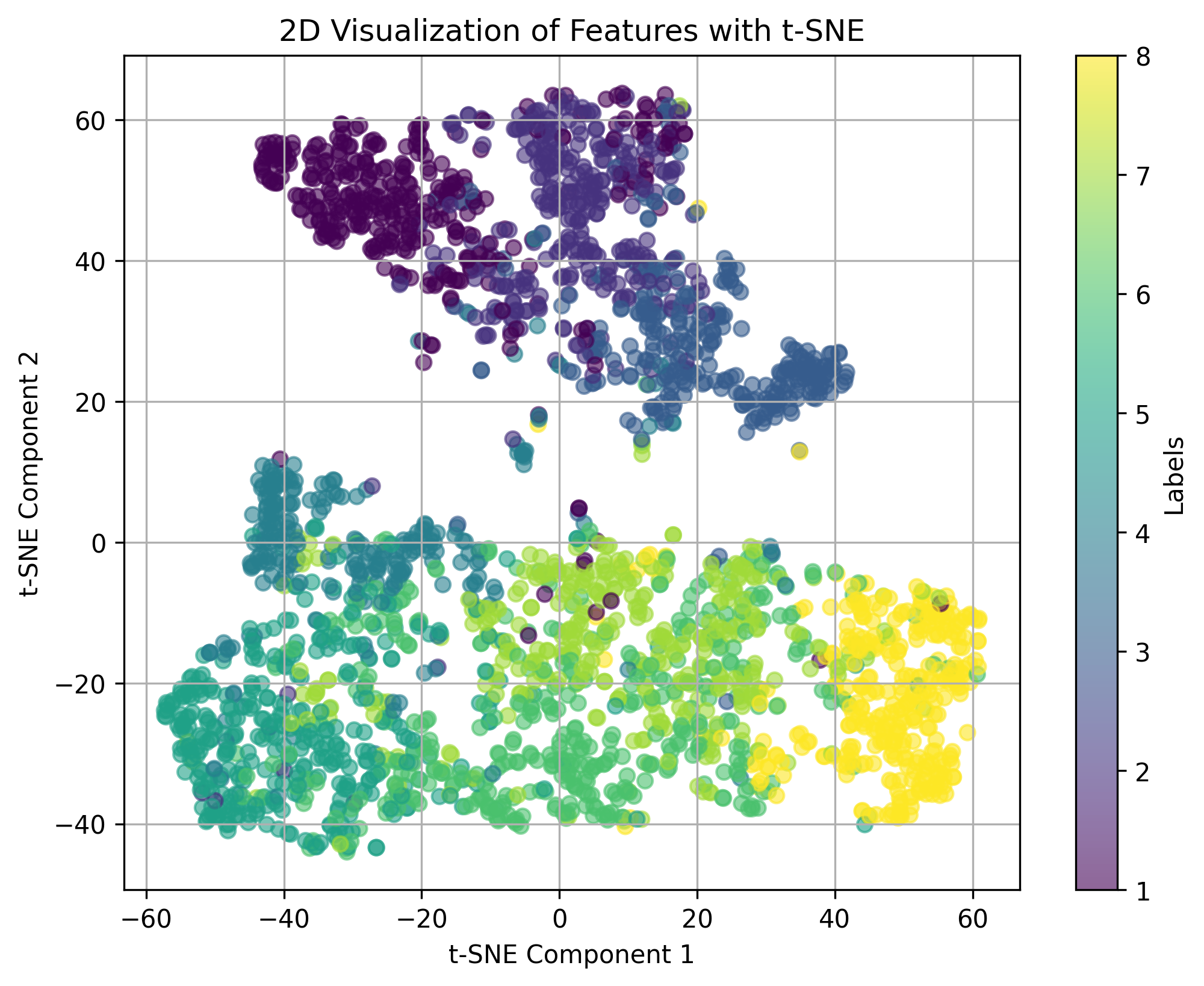}\label{fig:OutdoorScene_200}}
	\caption{The visualization results of the proposed method under 1, 50, 100 and 200 epochs on OutdoorScene dataset.}
	\label{fig:OutdoorScenee_tsne}
\end{figure*}

\textbf{Parameter sensitivity analysis}:
Our proposed BMvC method consists of a key balance parameter $\lambda$ to trade-off the feature reconstruction and view-specific contrastive regularization loss. To study the parameter sensitivity, we give the clustering performance measured by ACC, NMI, ARI, and Fscore on four datasets varying with different $\lambda$ in Fig. \ref{fig:ps_ap}, and more results are given in the appendix. From the results, we observe that the clustering performance of the proposed method slightly fluctuates with the $\lambda$. Additionally, when the parameter $\lambda$ is in the range $[10^-1, 10^0, 10^1]$, the proposed method obtains considerable clustering results. Therefore, we suggest setting the parameter $\lambda$ in range $[10^-1, 10^0, 10^1]$, when it is applied for some applications.

\textbf{Convergence analysis and visualization}:
In this part, we report the clustering performance and loss function values varying from training epochs on different datasets in Fig.~\ref{fig:loss}. The proposed method ensures good convergence properties. Additionally, we visualize the data distribution of the view-common feature presentation under different training epochs in Figs.~\ref{fig:CUB_tsne},~\ref{fig:MSRCV_tsne},~\ref{fig:nuswide_tsne}, and~\ref{fig:OutdoorScenee_tsne}. From the results, we find that the proposed method can reveal more compact and clear clustering results with the training epochs increase.

\section{Conclusions}
In this paper, we proposed a novel incomplete multi-view clustering method equipped with a mask-informed fusion strategy and prior knowledge-assisted contrastive learning.  The mask-informed fusion strategy flexibly aggregates incomplete multi-view information by considering the data observation status across multiple views, reducing the adverse impacts of missing values.  The prior knowledge-assisted contrastive learning leverages the neighbor information among different views to enhance the view-common feature representation with a weighted contrastive loss. Extensive experiments on various benchmark multi-view clustering datasets on both view complete and incomplete cases were conducted to verify the efficacy of our method.

The method demonstrates a slight advantage over others at missing rates of 0.5. This degradation occurs because views with high miss ratios make it challenging to construct accurate neighborhoods for the proposed prior knowledge-assisted contrastive loss, leading to performance degradation. In contrast, at missing rates between 0 and 0.3, our method shows significant advantages, indicating its suitability for incomplete multi-view datasets with lower missing rates. In future research, we propose to expand our approach to better address challenges associated with extreme missing rates.

\ifCLASSOPTIONcompsoc
\section*{Acknowledgments}
\else
\section*{Acknowledgment}
\fi

The authors wish to gratefully acknowledge the anonymous reviewers for the constructive comments of this paper.

\ifCLASSOPTIONcaptionsoff
\newpage
\fi

\bibliographystyle{IEEEtran}
\bibliography{IEEEabrv,References}

\begin{thebibliography}{10}
\providecommand{\url}[1]{#1}
\csname url@samestyle\endcsname
\providecommand{\newblock}{\relax}
\providecommand{\bibinfo}[2]{#2}
\providecommand{\BIBentrySTDinterwordspacing}{\spaceskip=0pt\relax}
\providecommand{\BIBentryALTinterwordstretchfactor}{4}
\providecommand{\BIBentryALTinterwordspacing}{\spaceskip=\fontdimen2\font plus
\BIBentryALTinterwordstretchfactor\fontdimen3\font minus
  \fontdimen4\font\relax}
\providecommand{\BIBforeignlanguage}[2]{{%
\expandafter\ifx\csname l@#1\endcsname\relax
\typeout{** WARNING: IEEEtran.bst: No hyphenation pattern has been}%
\typeout{** loaded for the language `#1'. Using the pattern for}%
\typeout{** the default language instead.}%
\else
\language=\csname l@#1\endcsname
\fi
#2}}
\providecommand{\BIBdecl}{\relax}
\BIBdecl

\bibitem{zhou2024survey}
L.~Zhou, G.~Du, K.~L{\"u}, L.~Wang, and J.~Du, ``A survey and an empirical
  evaluation of multi-view clustering approaches,'' \emph{ACM Computing
  Surveys}, vol.~56, no.~7, pp. 1--38, 2024.

\bibitem{jiang2022tensorial}
G.~Jiang, J.~Peng, H.~Wang, Z.~Mi, and X.~Fu, ``Tensorial multi-view clustering
  via low-rank constrained high-order graph learning,'' \emph{IEEE Transactions
  on Circuits and Systems for Video Technology}, vol.~32, no.~8, pp.
  5307--5318, 2022.

\bibitem{jia2021multi}
Y.~Jia, H.~Liu, J.~Hou, S.~Kwong, and Q.~Zhang, ``Multi-view spectral
  clustering tailored tensor low-rank representation,'' \emph{IEEE Transactions
  on Circuits and Systems for Video Technology}, vol.~31, no.~12, pp.
  4784--4797, 2021.

\bibitem{li2024mvbench}
K.~Li, Y.~Wang, Y.~He, Y.~Li, Y.~Wang, Y.~Liu, Z.~Wang, J.~Xu, G.~Chen, P.~Luo
  \emph{et~al.}, ``Mvbench: A comprehensive multi-modal video understanding
  benchmark,'' in \emph{IEEE Conference on Computer Vision and Pattern
  Recognition}, 2024, pp. 22\,195--22\,206.

\bibitem{schiappa2023self}
M.~C. Schiappa, Y.~S. Rawat, and M.~Shah, ``Self-supervised learning for
  videos: A survey,'' \emph{ACM Computing Surveys}, vol.~55, no. 13s, pp.
  1--37, 2023.

\bibitem{ke2023clustering}
G.~Ke, G.~Chao, X.~Wang, C.~Xu, Y.~Zhu, and Y.~Yu, ``A clustering-guided
  contrastive fusion for multi-view representation learning,'' \emph{IEEE
  Transactions on Circuits and Systems for Video Technology}, vol.~34, no.~4,
  pp. 2056--2069, 2023.

\bibitem{zhao2025scalable}
X.~Zhao, J.~Fan, X.~Chang, F.~Nie, Q.~Zhang, and J.~Guo, ``Scalable multi-view
  regression clustering for large-scale data,'' \emph{IEEE Transactions on
  Circuits and Systems for Video Technology}, 2025.

\bibitem{yan2024deep}
W.~Yan, K.~Liu, W.~Zhou, and C.~Tang, ``Deep incomplete multi-view clustering
  via dynamic imputation and triple alignment with dual optimization,''
  \emph{IEEE Transactions on Circuits and Systems for Video Technology}, 2024.

\bibitem{wen2023graph}
J.~Wen, G.~Xu, Z.~Tang, W.~Wang, L.~Fei, and Y.~Xu, ``Graph regularized and
  feature aware matrix factorization for robust incomplete multi-view
  clustering,'' \emph{IEEE Transactions on Circuits and Systems for Video
  Technology}, vol.~34, no.~5, pp. 3728--3741, 2024.

\bibitem{lin2021completer}
Y.~Lin, Y.~Gou, Z.~Liu, B.~Li, J.~Lv, and X.~Peng, ``Completer: Incomplete
  multi-view clustering via contrastive prediction,'' in \emph{IEEE Conference
  on Computer Vision and Pattern Recognition}, 2021, pp. 11\,174--11\,183.

\bibitem{li2023incomplete}
H.~Li, Y.~Li, M.~Yang, P.~Hu, D.~Peng, and X.~Peng, ``Incomplete multi-view
  clustering via prototype-based imputation,'' in \emph{International Joint
  Conference on Artificial Intelligence}, 2023, pp. 3911--3919.

\bibitem{pu2024adaptive}
J.~Pu, C.~Cui, X.~Chen, Y.~Ren, X.~Pu, Z.~Hao, S.~Y. Philip, and L.~He,
  ``Adaptive feature imputation with latent graph for deep incomplete
  multi-view clustering,'' in \emph{AAAI Conference on Artificial
  Intelligence}, vol.~38, no.~13, 2024, pp. 14\,633--14\,641.

\bibitem{meilua2024manifold}
M.~Meil{\u{a}} and H.~Zhang, ``Manifold learning: What, how, and why,''
  \emph{Annual Review of Statistics and Its Application}, vol.~11, 2024.

\bibitem{wen2020adaptive}
J.~Wen, K.~Yan, Z.~Zhang, Y.~Xu, J.~Wang, L.~Fei, and B.~Zhang, ``Adaptive
  graph completion based incomplete multi-view clustering,'' \emph{IEEE
  Transactions on Multimedia}, vol.~23, pp. 2493--2504, 2020.

\bibitem{yin2021incomplete}
J.~Yin and S.~Sun, ``Incomplete multi-view clustering with reconstructed
  views,'' \emph{IEEE Transactions on Knowledge and Data Engineering}, vol.~35,
  no.~3, pp. 2671--2682, 2021.

\bibitem{wang2022highly}
S.~Wang, X.~Liu, L.~Liu, W.~Tu, X.~Zhu, J.~Liu, S.~Zhou, and E.~Zhu,
  ``Highly-efficient incomplete large-scale multi-view clustering with
  consensus bipartite graph,'' in \emph{IEEE/CVF Conference on Computer Vision
  and Pattern Recognition}, 2022, pp. 9776--9785.

\bibitem{yang2022robust}
M.~Yang, Y.~Li, P.~Hu, J.~Bai, J.~Lv, and X.~Peng, ``Robust multi-view
  clustering with incomplete information,'' \emph{IEEE Transactions on Pattern
  Analysis and Machine Intelligence}, vol.~45, no.~1, pp. 1055--1069, 2022.

\bibitem{xu2024deep}
G.~Xu, J.~Wen, C.~Liu, B.~Hu, Y.~Liu, L.~Fei, and W.~Wang, ``Deep variational
  incomplete multi-view clustering: Exploring shared clustering structures,''
  in \emph{AAAI Conference on Artificial Intelligence}, vol.~38, no.~14, 2024,
  pp. 16\,147--16\,155.

\bibitem{hu2019one}
M.~Hu and S.~Chen, ``One-pass incomplete multi-view clustering,'' in \emph{AAAI
  Conference on Artificial Intelligence}, vol.~33, no.~01, 2019, pp.
  3838--3845.

\bibitem{khan2024weighted}
G.~A. Khan, J.~Khan, T.~Anwar, Z.~Ashraf, M.~H. Javed, and B.~Diallo,
  ``Weighted concept factorization based incomplete multi-view clustering,''
  \emph{IEEE Transactions on Artificial Intelligence}, 2024.

\bibitem{9556554}
X.~Liu, ``Incomplete multiple kernel alignment maximization for clustering,''
  \emph{IEEE Transactions on Pattern Analysis and Machine Intelligence},
  vol.~46, no.~3, pp. 1412--1424, 2024.

\bibitem{LI2024102086}
A.~Li, C.~Feng, Y.~Cheng, Y.~Zhang, and H.~Yang, ``Incomplete multiview
  subspace clustering based on multiple kernel low-redundant representation
  learning,'' \emph{Information Fusion}, vol. 103, p. 102086, 2024.

\bibitem{yang2024geometric}
Z.~Yang, H.~Zhang, Y.~Wei, Z.~Wang, F.~Nie, and D.~Hu, ``Geometric-inspired
  graph-based incomplete multi-view clustering,'' \emph{Pattern Recognition},
  vol. 147, p. 110082, 2024.

\bibitem{du2024fast}
L.~Du, Y.~Shi, Y.~Chen, P.~Zhou, and Y.~Qian, ``Fast and scalable incomplete
  multi-view clustering with duality optimal graph filtering,'' in \emph{ACM
  International Conference on Multimedia}, 2024, pp. 8893--8902.

\bibitem{li2022refining}
X.-L. Li, M.-S. Chen, C.-D. Wang, and J.-H. Lai, ``Refining graph structure for
  incomplete multi-view clustering,'' \emph{IEEE Transactions on Neural
  Networks and Learning Systems}, vol.~35, no.~2, pp. 2300--2313, 2022.

\bibitem{elhamifar2013sparse}
E.~Elhamifar and R.~Vidal, ``Sparse subspace clustering: Algorithm, theory, and
  applications,'' \emph{IEEE Transactions on Pattern Analysis and Machine
  Intelligence}, vol.~35, no.~11, pp. 2765--2781, 2013.

\bibitem{nie2014clustering}
F.~Nie, X.~Wang, and H.~Huang, ``Clustering and projected clustering with
  adaptive neighbors,'' in \emph{International Conference on Knowledge
  Discovery and Data Mining}, 2014, pp. 977--986.

\bibitem{wen2020dimc}
J.~Wen, Z.~Zhang, Z.~Zhang, Z.~Wu, L.~Fei, Y.~Xu, and B.~Zhang, ``Dimc-net:
  Deep incomplete multi-view clustering network,'' in \emph{ACM International
  Conference on Multimedia}, 2020, pp. 3753--3761.

\bibitem{xu2022deep}
J.~Xu, C.~Li, Y.~Ren, L.~Peng, Y.~Mo, X.~Shi, and X.~Zhu, ``Deep incomplete
  multi-view clustering via mining cluster complementarity,'' in \emph{AAAI
  Conference on Artificial Intelligence}, vol.~36, no.~8, 2022, pp. 8761--8769.

\bibitem{xue2024robust}
Z.~Xue, Y.~Li, Z.~Guan, W.~Li, M.~Liang, and H.~Zhou, ``Robust multi-graph
  contrastive network for incomplete multi-view clustering,'' \emph{IEEE
  Transactions on Multimedia}, 2024.

\bibitem{krishnan2022self}
R.~Krishnan, P.~Rajpurkar, and E.~J. Topol, ``Self-supervised learning in
  medicine and healthcare,'' \emph{Nature Biomedical Engineering}, vol.~6,
  no.~12, pp. 1346--1352, 2022.

\bibitem{liu2021self}
X.~Liu, F.~Zhang, Z.~Hou, L.~Mian, Z.~Wang, J.~Zhang, and J.~Tang,
  ``Self-supervised learning: Generative or contrastive,'' \emph{IEEE
  Transactions on Knowledge and Data Engineering}, vol.~35, no.~1, pp.
  857--876, 2021.

\bibitem{khosla2020supervised}
P.~Khosla, P.~Teterwak, C.~Wang, A.~Sarna, Y.~Tian, P.~Isola, A.~Maschinot,
  C.~Liu, and D.~Krishnan, ``Supervised contrastive learning,'' \emph{Advances
  in Neural Information Processing Systems}, vol.~33, pp. 18\,661--18\,673,
  2020.

\bibitem{he2020momentum}
K.~He, H.~Fan, Y.~Wu, S.~Xie, and R.~Girshick, ``Momentum contrast for
  unsupervised visual representation learning,'' in \emph{IEEE/CVF conference
  on Computer Vision and Pattern Recognition}, 2020, pp. 9729--9738.

\bibitem{chen2020simple}
T.~Chen, S.~Kornblith, M.~Norouzi, and G.~Hinton, ``A simple framework for
  contrastive learning of visual representations,'' in \emph{International
  Conference on Machine Learning}.\hskip 1em plus 0.5em minus 0.4em\relax PMLR,
  2020, pp. 1597--1607.

\bibitem{caron2020unsupervised}
M.~Caron, I.~Misra, J.~Mairal, P.~Goyal, P.~Bojanowski, and A.~Joulin,
  ``Unsupervised learning of visual features by contrasting cluster
  assignments,'' \emph{Advances in Neural Information Processing Systems},
  vol.~33, pp. 9912--9924, 2020.

\bibitem{gui2024survey}
J.~Gui, T.~Chen, J.~Zhang, Q.~Cao, Z.~Sun, H.~Luo, and D.~Tao, ``A survey on
  self-supervised learning: Algorithms, applications, and future trends,''
  \emph{IEEE Transactions on Pattern Analysis and Machine Intelligence}, 2024.

\bibitem{xu2022multi}
J.~Xu, H.~Tang, Y.~Ren, L.~Peng, X.~Zhu, and L.~He, ``Multi-level feature
  learning for contrastive multi-view clustering,'' in \emph{IEEE Conference on
  Computer Vision and Pattern Recognition}, 2022, pp. 16\,051--16\,060.

\bibitem{trosten2023effects}
D.~J. Trosten, S.~L{\o}kse, R.~Jenssen, and M.~C. Kampffmeyer, ``On the effects
  of self-supervision and contrastive alignment in deep multi-view
  clustering,'' in \emph{IEEE Conference on Computer Vision and Pattern
  Recognition}, 2023, pp. 23\,976--23\,985.

\bibitem{luo2024simple}
C.~Luo, J.~Xu, Y.~Ren, J.~Ma, and X.~Zhu, ``Simple contrastive multi-view
  clustering with data-level fusion,'' in \emph{International Joint Conference
  on Artificial Intelligence}, 2024, pp. 4697--4705.

\bibitem{xia2021graph}
F.~Xia, K.~Sun, S.~Yu, A.~Aziz, L.~Wan, S.~Pan, and H.~Liu, ``Graph learning: A
  survey,'' \emph{IEEE Transactions on Artificial Intelligence}, vol.~2, no.~2,
  pp. 109--127, 2021.

\bibitem{he2003locality}
X.~He and P.~Niyogi, ``Locality preserving projections,'' \emph{Advances in
  Neural Information Processing Systems}, vol.~16, 2003.

\bibitem{gander1980algorithms}
W.~Gander, ``Algorithms for the qr decomposition,'' \emph{Res. Rep}, vol.~80,
  no.~02, pp. 1251--1268, 1980.

\bibitem{wang2021understanding}
F.~Wang and H.~Liu, ``Understanding the behaviour of contrastive loss,'' in
  \emph{IEEE Conference on Computer Vision and Pattern Recognition}, 2021, pp.
  2495--2504.

\bibitem{yeh2022decoupled}
C.-H. Yeh, C.-Y. Hong, Y.-C. Hsu, T.-L. Liu, Y.~Chen, and Y.~LeCun, ``Decoupled
  contrastive learning,'' in \emph{European Conference on Computer
  Vision}.\hskip 1em plus 0.5em minus 0.4em\relax Springer, 2022, pp. 668--684.

\bibitem{hu2020multi}
Z.~Hu, F.~Nie, R.~Wang, and X.~Li, ``Multi-view spectral clustering via
  integrating nonnegative embedding and spectral embedding,'' \emph{Information
  Fusion}, vol.~55, pp. 251--259, 2020.

\bibitem{zhen2019deep}
L.~Zhen, P.~Hu, X.~Wang, and D.~Peng, ``Deep supervised cross-modal
  retrieval,'' in \emph{IEEE Conference on Computer Vision and Pattern
  Recognition}, 2019, pp. 10\,394--10\,403.

\bibitem{paszke2019pytorch}
A.~Paszke, S.~Gross, F.~Massa, A.~Lerer, J.~Bradbury, G.~Chanan, T.~Killeen,
  Z.~Lin, N.~Gimelshein, L.~Antiga \emph{et~al.}, ``Pytorch: An imperative
  style, high-performance deep learning library,'' \emph{Advances in Neural
  Information Processing Systems}, vol.~32, 2019.

\bibitem{kingma2014adam}
D.~P. Kingma, ``Adam: A method for stochastic optimization,'' \emph{arXiv
  preprint arXiv:1412.6980}, 2014.

\bibitem{wang2015deep}
W.~Wang, R.~Arora, K.~Livescu, and J.~Bilmes, ``On deep multi-view
  representation learning,'' in \emph{International Conference on Machine
  Learning}.\hskip 1em plus 0.5em minus 0.4em\relax PMLR, 2015, pp. 1083--1092.

\bibitem{nie2018multiview}
F.~Nie, L.~Tian, and X.~Li, ``Multiview clustering via adaptively weighted
  procrustes,'' in \emph{International Conference on Knowledge Discovery and
  Data Mining}, 2018, pp. 2022--2030.

\bibitem{wang2019gmc}
H.~Wang, Y.~Yang, and B.~Liu, ``Gmc: Graph-based multi-view clustering,''
  \emph{IEEE Transactions on Knowledge and Data Engineering}, vol.~32, no.~6,
  pp. 1116--1129, 2020.

\bibitem{liu2021one}
J.~Liu, X.~Liu, Y.~Yang, L.~Liu, S.~Wang, W.~Liang, and J.~Shi, ``One-pass
  multi-view clustering for large-scale data,'' in \emph{International
  Conference on Computer Vision}, 2021, pp. 12\,344--12\,353.

\bibitem{chen2023deep}
J.~Chen, H.~Mao, W.~L. Woo, and X.~Peng, ``Deep multiview clustering by
  contrasting cluster assignments,'' in \emph{IEEE Conference on Computer
  Vision and Pattern Recognition}, 2023, pp. 16\,752--16\,761.

\bibitem{wang2023efficient}
J.~Wang, C.~Tang, Z.~Wan, W.~Zhang, K.~Sun, and A.~Y. Zomaya, ``Efficient and
  effective one-step multiview clustering,'' \emph{IEEE Transactions on Neural
  Networks and Learning Systems}, 2023.

\bibitem{wu2024self}
S.~Wu, Y.~Zheng, Y.~Ren, J.~He, X.~Pu, S.~Huang, Z.~Hao, and L.~He,
  ``Self-weighted contrastive fusion for deep multi-view clustering,''
  \emph{IEEE Transactions on Multimedia}, 2024.

\bibitem{tang2022deep}
H.~Tang and Y.~Liu, ``Deep safe incomplete multi-view clustering: Theorem and
  algorithm,'' in \emph{International Conference on Machine Learning}.\hskip
  1em plus 0.5em minus 0.4em\relax PMLR, 2022, pp. 21\,090--21\,110.

\bibitem{Yan_2023_CVPR}
W.~Yan, Y.~Zhang, C.~Lv, C.~Tang, G.~Yue, L.~Liao, and W.~Lin, ``Gcfagg: Global
  and cross-view feature aggregation for multi-view clustering,'' in
  \emph{IEEE/CVF Conference on Computer Vision and Pattern Recognition}, June
  2023, pp. 19\,863--19\,872.

\bibitem{jin2023deep}
J.~Jin, S.~Wang, Z.~Dong, X.~Liu, and E.~Zhu, ``Deep incomplete multi-view
  clustering with cross-view partial sample and prototype alignment,'' in
  \emph{IEEE/CVF Conference on Computer Vision and Pattern Recognition}, 2023,
  pp. 11\,600--11\,609.

\bibitem{xu2024investigating}
J.~Xu, Y.~Ren, X.~Wang, L.~Feng, Z.~Zhang, G.~Niu, and X.~Zhu, ``Investigating
  and mitigating the side effects of noisy views for self-supervised clustering
  algorithms in practical multi-view scenarios,'' in \emph{IEEE Conference on
  Computer Vision and Pattern Recognition}, 2024, pp. 22\,957--22\,966.

\end{thebibliography}

\end{document}